\documentclass[10pt,twocolumn,letterpaper]{article}

\usepackage{icb}
\usepackage{times}
\usepackage{epsfig}
\usepackage{graphicx}
\usepackage{amsmath}
\usepackage{amssymb}
\usepackage[tight]{subfigure}
\usepackage{url}
\usepackage{enumitem}
\usepackage{multirow}
\usepackage[nodisplayskipstretch]{setspace}
\usepackage{algorithm}
\usepackage{color}
\usepackage[noend]{algpseudocode}

\icbfinalcopy % *** Uncomment this line for the final submission

\begin{document}

%%%%%%%%% TITLE
\title{ Latent Fingerprint Recognition: Role of Texture Template}

\author{Kai Cao and Anil K. Jain\\
Department of Computer Science and Engineering\\
Michigan State University, East Lansing, Michigan 48824\\
{\tt\small Email: \{kaicao,jain\}@cse.msu.edu} }

 \maketitle
\thispagestyle{empty}

%%%%%%%%% ABSTRACT
\begin{abstract}

%Virtual minutiae based texture templates (non-minutiae templates) are critical to improve  overall latent identification accuracy 

\end{abstract}
We propose a texture template approach, consisting of a set of virtual minutiae, to improve the overall latent fingerprint recognition accuracy.  To compensate for the lack of sufficient number of minutiae in poor quality latent prints, we generate a set of virtual minutiae.  However, due to a large number of these regularly placed virtual minutiae, texture based template matching has a  large computational requirement compared to matching true minutiae templates. To improve both the accuracy and efficiency of the texture template matching, we investigate: i) both original and enhanced fingerprint patches for training convolutional neural networks (ConvNets) to improve the distinctiveness of descriptors associated with each virtual minutiae, ii) smaller patches around virtual minutiae and a fast ConvNet architecture to speed up descriptor extraction, iii) reduce the descriptor length, iv) a modified hierarchical graph matching strategy to improve the matching speed, and v) extraction of multiple texture templates to boost the performance. Experiments on NIST SD27 latent database show that the above strategies can improve the matching speed from 11 ms (24 threads) per comparison (between a latent and a reference print)  to only 7.7 ms (single thread) per comparison while improving the rank-1 accuracy by 8.9\% against 10K gallery.  
%In order to improve the feature extraction speed, smaller patch sizes are used to feed to convolutional neural network (CNN) and larger stride is used to reduce the number of virtual minutiae. 

 %

%%%%%%%%% BODY TEXT
\section{Introduction}
\vspace*{-5pt}
\begin{figure}[t]
	\centering
	
	\subfigure[]{
		\includegraphics[clip, trim=2cm 1cm 2cm 1cm, width=0.45\linewidth]{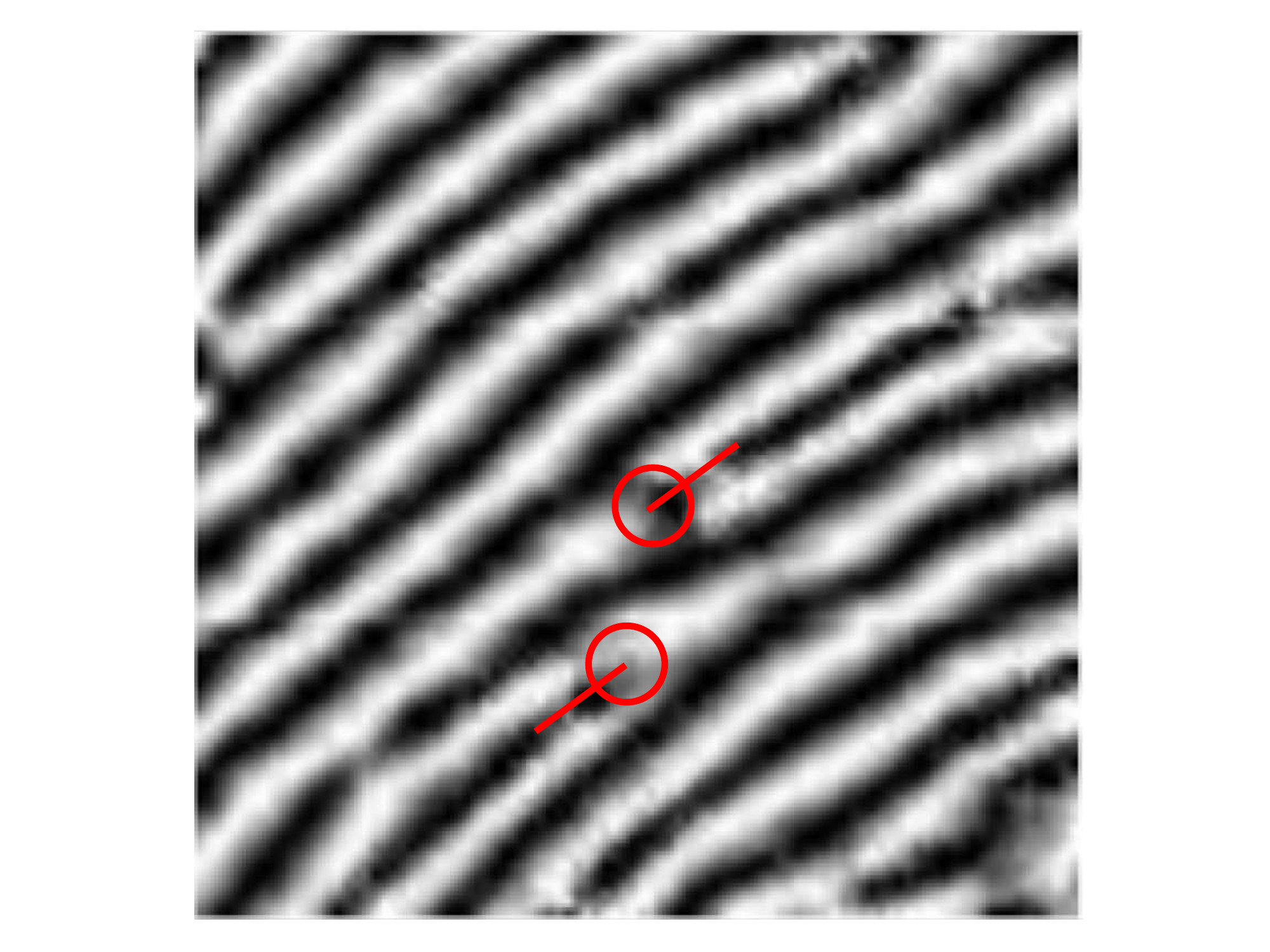}
	}
	\subfigure[]{
		\includegraphics[clip, trim=6.5cm 12cm 10cm 11.8cm, width=0.45\linewidth]{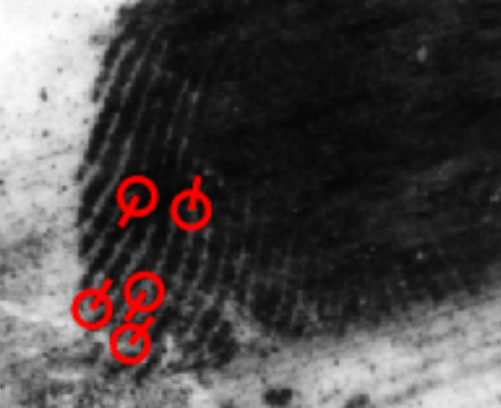}
	}           
	\hspace{0.2 cm}
	\subfigure[]{
		\includegraphics[clip, trim=4cm 8cm 4cm 8cm, width=0.55\linewidth]{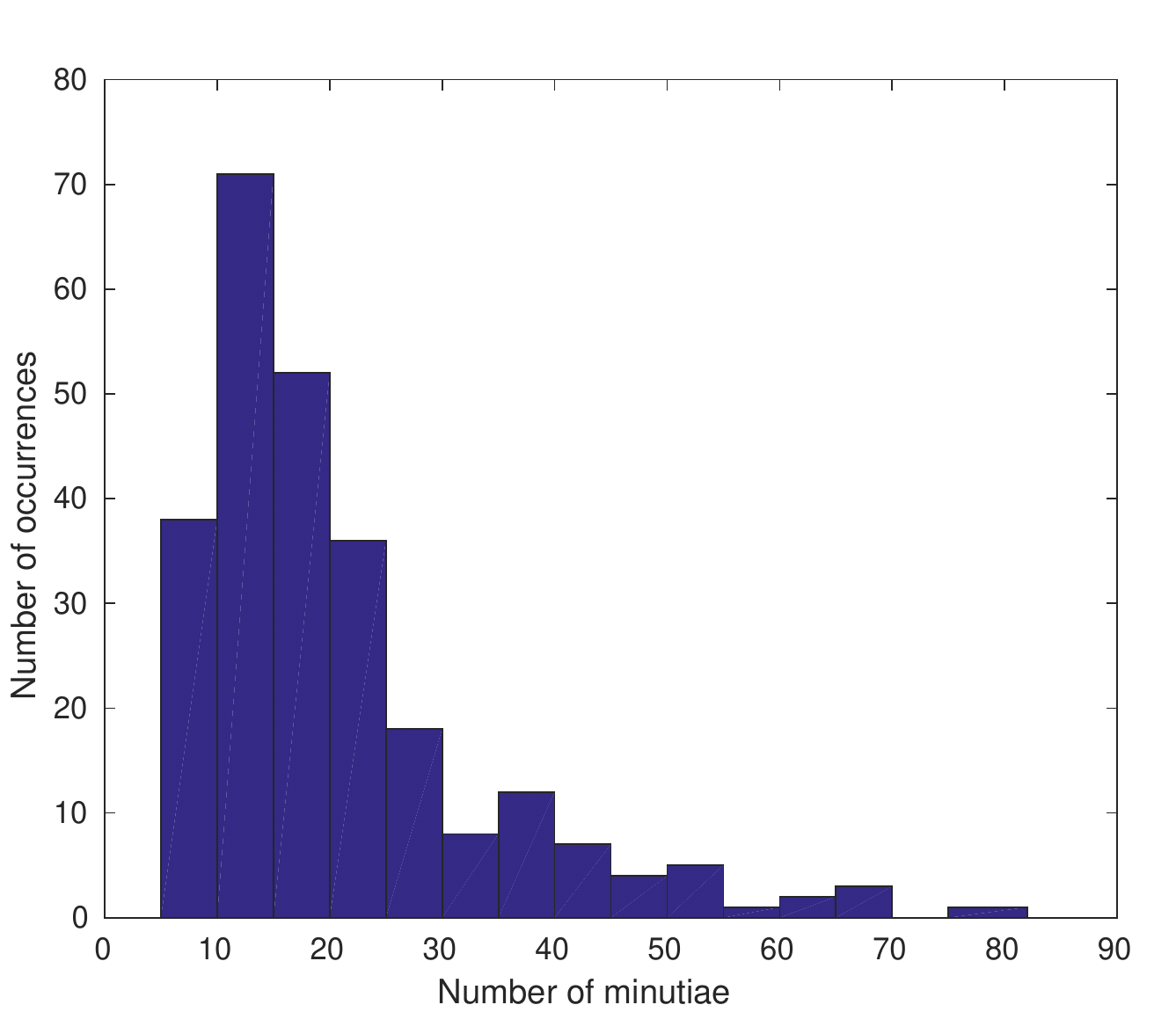}
	}
	\caption{Fingerprint texture for matching.  (a) Only two minutiae in a fingerprint image  (96$\times $ 96 at 500 ppi) acquired by a capacitive sensor embedded in a smartphone (provided by Goodix), (b) five manually marked minutiae in a latent fingerprint image from NIST SD27, and (c) histogram of the number of minutiae in latent fingerprints in NIST SD27 database (consisting of 258 operational latents).}
	\label{fig:introduction}
	\vspace*{-5pt}
\end{figure}

Ever since minutiae were introduced for comparing fingerprints in 1888 by Sir Francis Galton \cite{Galton},  minutiae are considered as the foundation for the science of fingerprint identification, which has expanded and transitioned to a wide variety of applications for person recognition over the past century \cite{FingerprintHandbook}.  The first Automated Fingerprint Identification System (AFIS)  launched by FBI in early 1970s only stored  type of fingerprint and its minutiae instead of digital images because of the compact and efficient representation offered by minutiae\footnote{https://www.fbi.gov/file-repository/about-us-cjis-fingerprints\_biometrics-biometric-center-of-excellences-fingerprint-recognition.pdf/view}.  With decades of research and development  and advances in processor, memory and sensor design,  fingerprint recognition systems have now been deployed in a broad set of applications such as border
control, employment background checks, secure facility access and national identity programs \cite{FingerprintHandbook}. %  not only in law enforcement and forensic agencies but also in numerous civilian applications.
 Aadhaar\footnote{https://uidai.gov.in/about-uidai/about-uidai.html},  has the world's largest biometric ID system with an enrollment database that already exceeds 1.2 billion tenprints
(along with corresponding irises and photos) of supposedly unique individuals. All of these systems are primarily based on minutiae based fingerprint matching algorithms. 

 % % % % % % ------

Minutiae based approaches, however, may not be effective in some cases, for example, in poor quality latent fingerprint matching and fingerprint images captured by small area sensors in mobile phones.  Latent fingerprints (latents)  are one of the
most important and widely used sources of evidence in law enforcement and forensic agencies worldwide \cite{Hawthorne2002}. Due to the unintentional deposition of the print
by a subject, latents are typically of poor quality in terms of ridge clarity, large background noise and small friction ridge area. Hence, the number of minutiae in a latent may be very small , e.g., $\leq$10. 
  Fig. \ref{fig:introduction} (c) shows the distribution of the number of manually marked minutiae on NIST SD27 latent database and  Fig. \ref{fig:introduction} (b) shows a latent image with 5 manually marked minutiae overlaid on the image. Minutiae alone do not have enough information for latents.
% were stored because the cost of storage for the digital images of the fingerprints was prohibitive.
 Another example where minutiae based matching does not work is for matching fingerprint images captured by the capacitive sensors embedded in smartphones.  It is estimated that 67\% of the smartphones in the world  will have an embedded fingerprint sensor\footnote{https://www.statista.com/statistics/522058/global-smartphone-fingerprint-penetration/} by 2018. The size of these embedded fingerprint sensors (only $88 \times 88$ pixels for TocuhID\footnote{https://assets.documentcloud.org/documents/1302613/ios-security-guide-sept-2014.pdf}) are much smaller than the standalone sensors. Hence, the number of minutiae in these images is very few as shown in Fig. \ref{fig:introduction} (a). For these reasons, accurate non-minutiae based (also called \textit{texture based})  fingerprint matching algorithms are necessary.  To our knowledge, all major latent AFIS vendors use texture templates and different quality latents are handled differently inside AFIS and that is why the processing time is different. 
%{https://www.fbi.gov/file-repository/about-us-cjis-fingerprints_biometrics-biometric-center-of-excellences-fingerprint-recognition.pdf}

A few non-minutiae based fingerprint matching algorithms have been proposed in literature.  FingerCode by Jain et al. \cite{Jain2000IP} uses a bank of Gabor filters to capture both the local and global  details in a fingerprint. However, FingerCode relies on a reference point and, further, its accuracy is much lower than minutiae based approaches.  Some approaches based on keypoints, e.g., SIFT \cite{Park, Yamazaki} and AKAZE \cite{AKAZE},  have been proposed to generate  dense keypoints for fingerprint matching.  But these keypoints as well as their descriptors are not sufficient to distinguish fingerprints from different fingers.

Deep learning based approaches have also been proposed for fingerprint recognition.  Zhang et al. \cite{Zhang2017IJCB} proposed a deep learning based feature, called deep dense multi-level feature, for partial high resolution fingerprint recognition which achieved promising performance on their own database. However, their approach could not handle fingerprint rotation. Cao and Jain \cite{Cao2018PAMI} proposed a virtual minutiae based approach for latent fingerprint recognition, where the virtual minutiae locations are determined by a raster scan with a stride of 16 pixels; the associated descriptors are obtained by three convolutional neural networks. Experimental results on two latent databases,  NIST SD27 and WVU, showed that the recognition performance of virtual minutiae based ``texture template" when fused with two different true minutiae templates boosts the rank-1 accuracy  from 58.5\% to 64.7\%  against 100K reference print gallery for NIST SD27 \cite{Cao2018PAMI}. However, the virtual minutiae feature extractor and matcher are quite slow. 

%, which is a bottle neck in the overall matching. We use patches from both the original fingerprints  and the enhanced fingerprint to: i) augment the training dataset and ii) improve descriptor distinctiveness. In order to improve the algorithm efficiency, we  use i) a faster CNN architecture, i.e, MobileNet,  ii) smaller patch size, $96\times 96$ pixels, to speed up descriptor extraction, iii) reduce the descriptor length, and iv)  improve the speed of the second-order graph matching.

The objective of this paper is to improve both the accuracy and efficiency of virtual minutiae (texture template) based latent matching. The main contributions of this paper are as follows: 
 \begin{enumerate} 
\itemsep0em 
\item Reduce  the average recognition time between a latent texture template and a rolled texture template  from 11 ms (24 threads) to 7.7 ms (single thread);
\item Improve the rank-1 identification rate of the texture templates in \cite{Cao2018PAMI} by 8.9\% (from 59.3\% to 68.2\%) for 10K gallery;
\item Boost the rank-1 identification rate  in \cite{Cao2018PAMI} by 2.7\% by fusion  of the proposed three texture templates with three templates in \cite{Cao2018PAMI} (from 75.6\% to 78.3\%) for 10K gallery. This means that out of the 258 latents in NIST SD27, improvements in the texture template will push 7 additional latents at rank 1.
 \end{enumerate}

\section{Proposed Approach}

\begin{figure*}[t]
	\centering
	\includegraphics[clip, trim=0cm 0cm 0cm 0cm, width=0.85\linewidth]{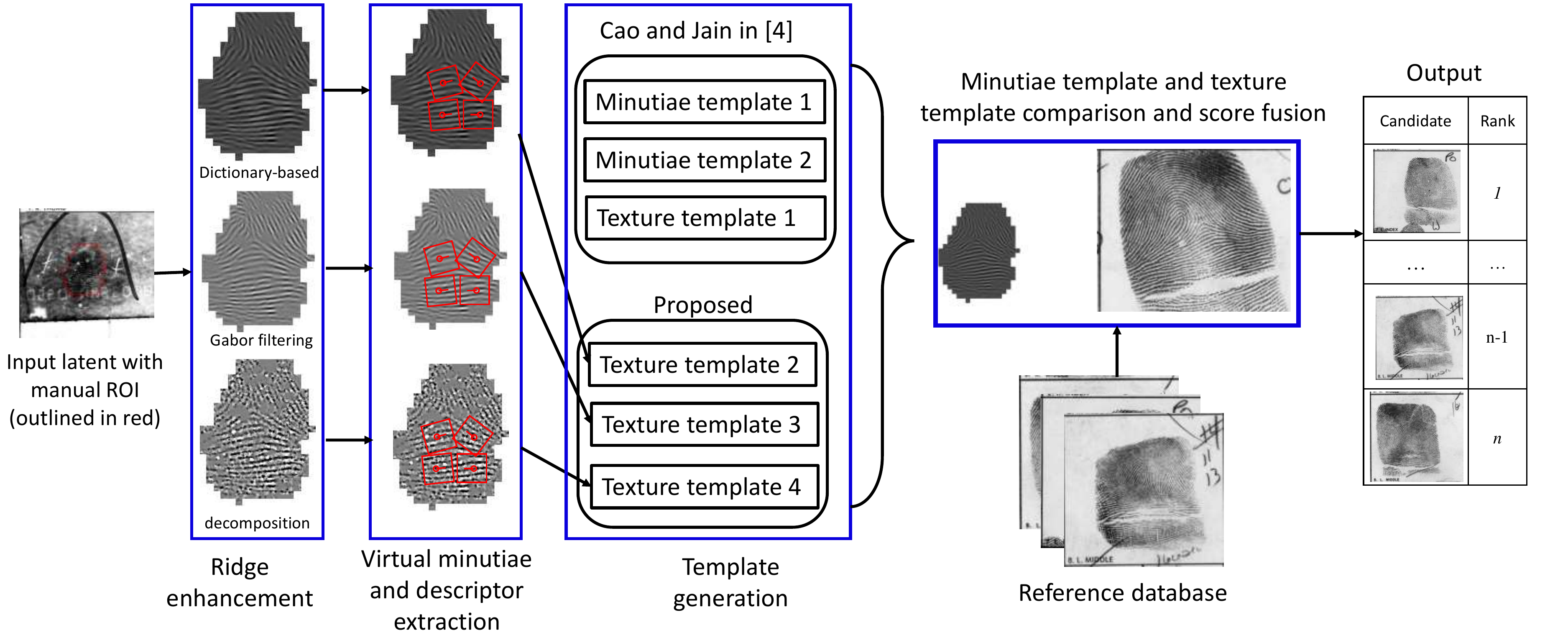}
	\caption{Overview of the proposed latent fingerprint recognition algorithm. }
	\label{fig:overview}
	\setlength{\belowdisplayskip}{-6pt}
	\setlength{\belowdisplayshortskip}{-6pt}
	\vspace*{-5pt}
\end{figure*}

In this section, we describe the proposed texture-based latent matching approach, including feature extraction and matching. Fig. \ref{fig:overview} shows flowchart of the proposed approach.

\subsection{Virtual Minutiae Extraction}
For both latents and reference prints, the texture template is similar to that in \cite{Cao2018PAMI} and consists of locations, orientations and descriptors of virtual minutiae.  We first describe the virtual minutiae localization and then discuss the associated descriptors. 

\begin{figure}[t]
	\centering
	\subfigure[]{
		\includegraphics[clip, trim=3cm 1cm 3cm 1cm, width=0.4\linewidth]{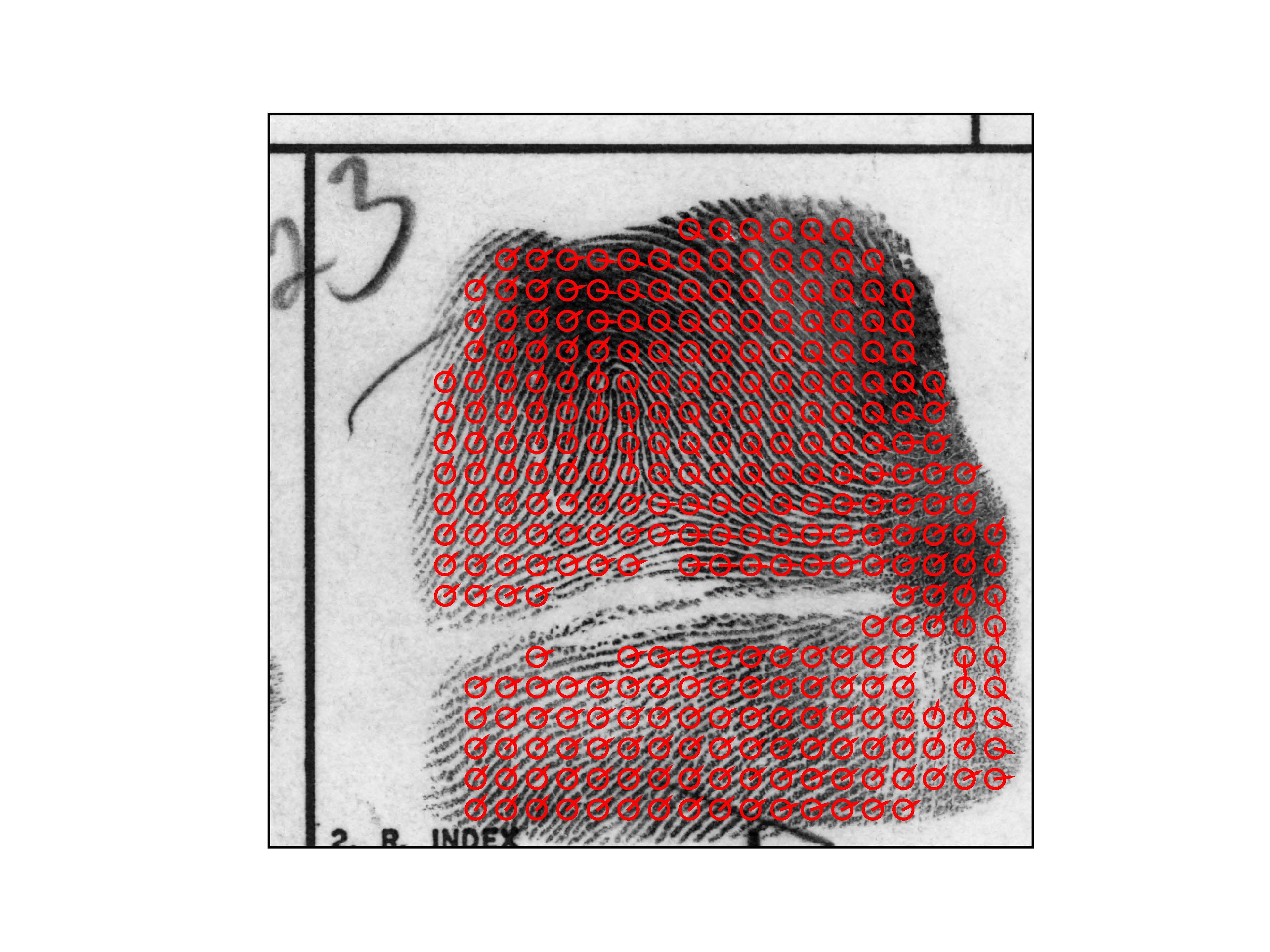}
	}\hspace{1cm}
	\subfigure[]{
		\includegraphics[clip, trim=3cm 1cm 3cm 1cm, width=0.4\linewidth]{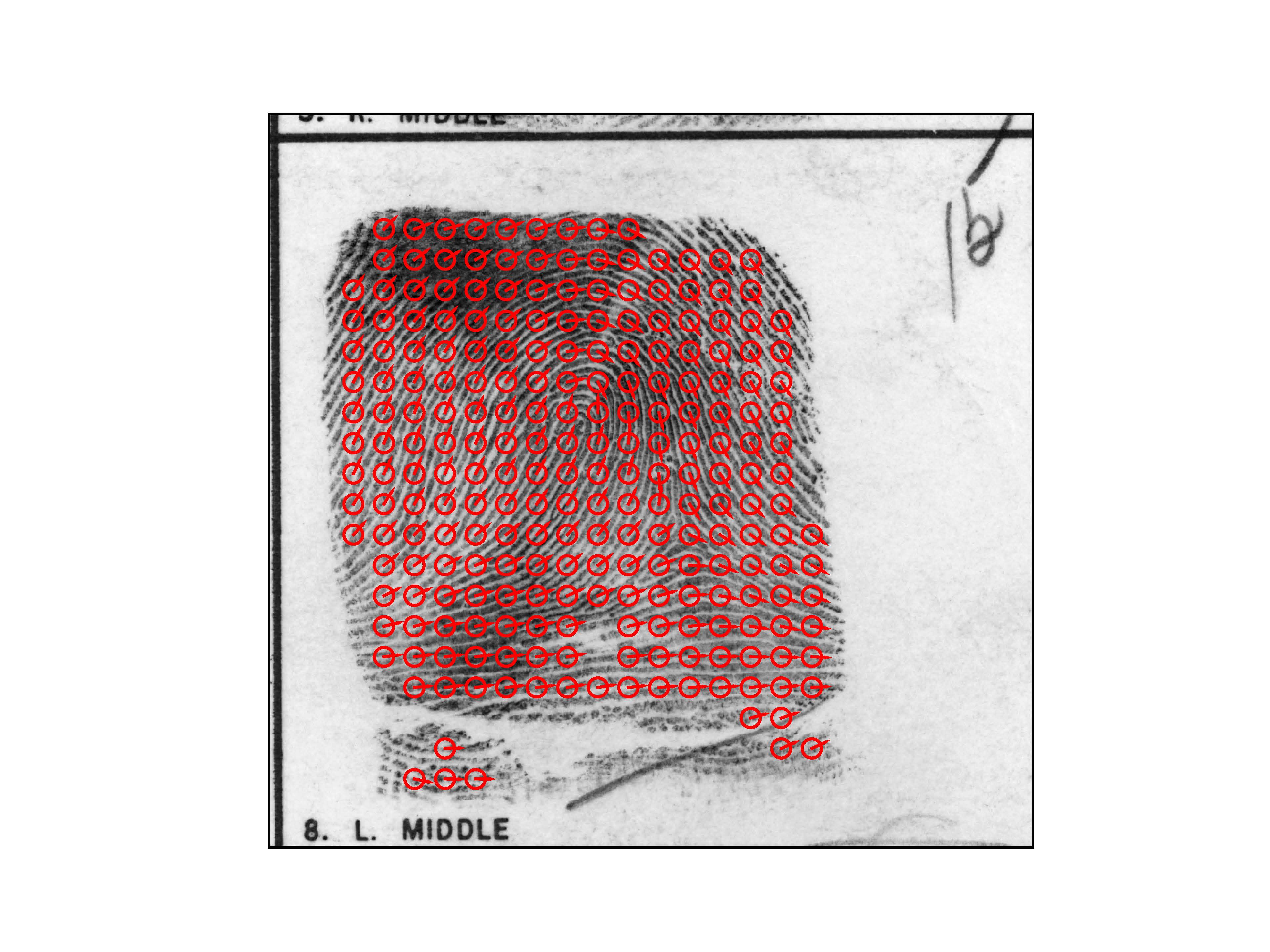}
	}           
	\caption{Virtual minutiae on two rolled prints with stride $s$=32.  }
	\label{fig:rolled_virtual}
	\vspace*{-5pt}
\end{figure}

For reference fingerprints which are typically of good quality, the region of interest (ROI) is   segmented by magnitude of the gradient and  the orientation fields with a block size of $16\times 16$ pixels as in \cite{Chikkerur2007198}. %Fig. \ref{fig:reference_print} (a) and (b) shows the orientation fields within the ROIs of two example reference prints. 
The locations of virtual minutiae are sampled by raster scan with a stride of $s$ and their orientations are the same as the orientations of their nearest blocks in the orientation field. The virtual minutiae close to the mask border are ignored.  Fig. \ref{fig:rolled_virtual} shows the virtual minutiae  on two rolled prints. % in Fig. \ref{fig:reference_print} (a).

\begin{figure}[t]
	\centering
	\subfigure[]{
		\includegraphics[clip, trim=5cm 5.5cm 6cm 1.5cm, width=0.35\linewidth]{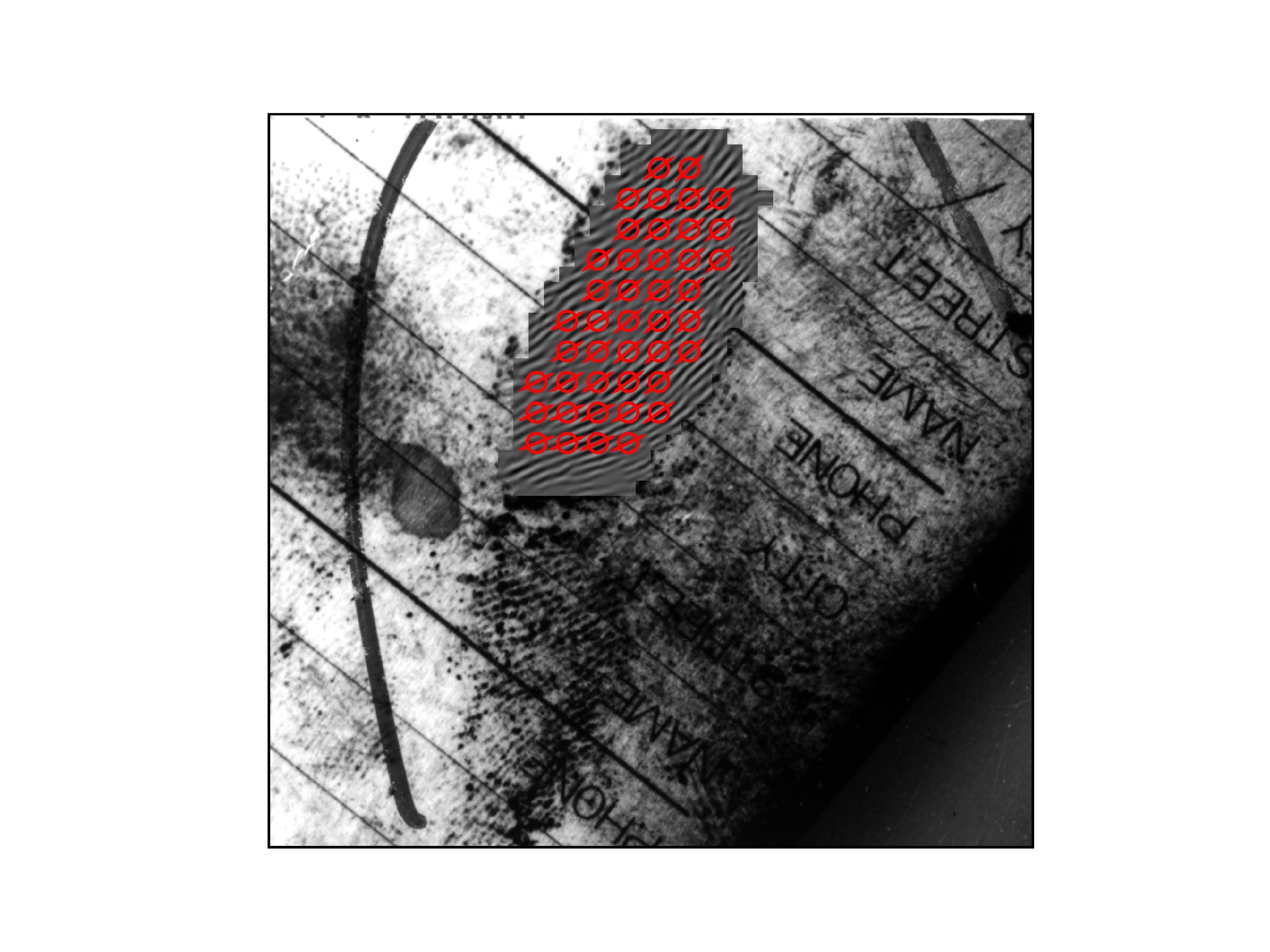}
	}
\hspace{1cm}
	\subfigure[]{
		\includegraphics[clip, trim=5cm 3.5cm 6cm 3.5cm, width=0.35\linewidth]{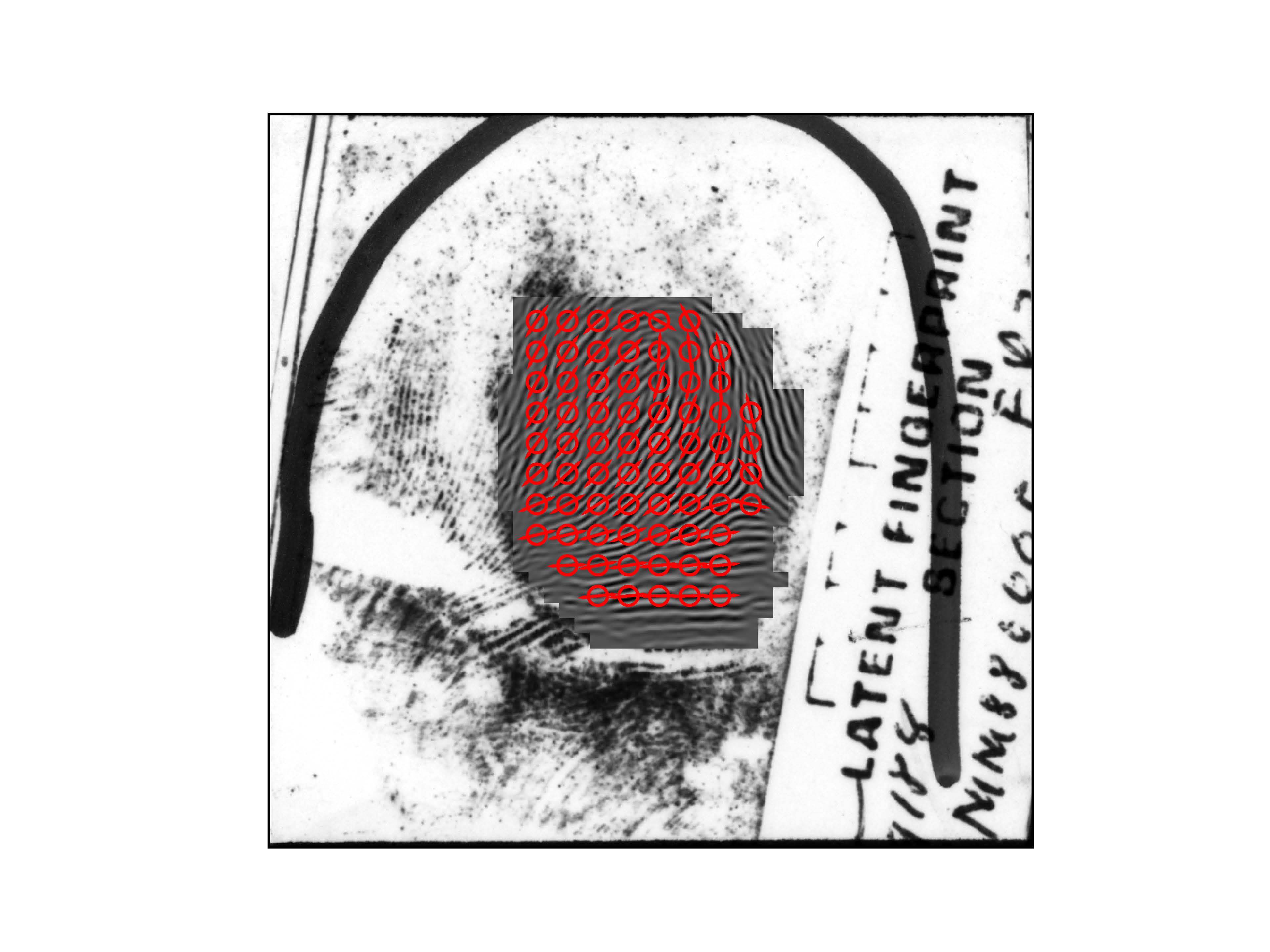}
	}           
	\caption{Virtual minutiae on two enhanced latent fingerprints with stride $s$=32. Note that each circle represents two virtual minutiae with opposite orientations to handle the ambiguity in ridge orientation.}
	\label{fig:latent_virtual}
	\vspace*{-5pt}
\end{figure}

For latents, the same manually marked ROIs and automatically
extracted ridge flow with a block size of $16\times 16$ pixels as in \cite{Cao2015ICB} are used for
virtual minutiae extraction.  Suppose that $(x, y)$ are the $x-$ and $y-$coordinates of a sampling point and $\theta$ denotes the orientation of the block which is closest to $(x,y)$ in the ridge flow. Two  virtual minutiae, i.e., $(x,y,\theta)$  and $(x,y,\theta+\pi)$, are created to handle the ambiguity in ridge orientation.  Fig. \ref{fig:latent_virtual} show virtual minutiae
on two enhanced latents from NIST SD27 with $s = 32$.

\subsection{Descriptors for Virtual Minutiae}

\begin{figure}[t]
	\centering
	\subfigure[][]{
		\includegraphics[clip, trim=0cm 0cm 0cm 0cm, width=0.21\linewidth]{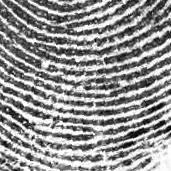}
	}
	\subfigure[]{
		\includegraphics[clip, trim=0cm 0cm 0cm 0cm, width=0.21\linewidth]{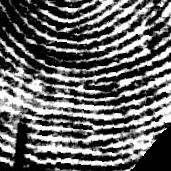}
	}     
	\subfigure[]{
		\includegraphics[clip, trim=0cm 0cm 0cm 0cm, width=0.21\linewidth]{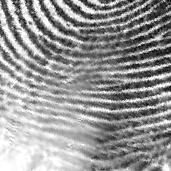}
	}    
	\subfigure[]{
		\includegraphics[clip, trim=0cm 0cm 0cm 0cm, width=0.21\linewidth]{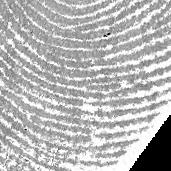}
	}    
	\subfigure[]{
		\includegraphics[clip, trim=0cm 0cm 0cm 0cm, width=0.21\linewidth]{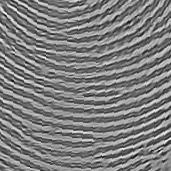}
	}
	\subfigure[]{
		\includegraphics[clip, trim=0cm 0cm 0cm 0cm, width=0.21\linewidth]{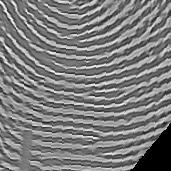}
	}     
	\subfigure[]{
		\includegraphics[clip, trim=0cm 0cm 0cm 0cm, width=0.21\linewidth]{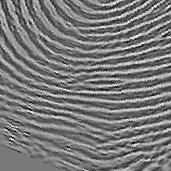}
	}    
	\subfigure[]{
		\includegraphics[clip, trim=0cm 0cm 0cm 0cm, width=0.21\linewidth]{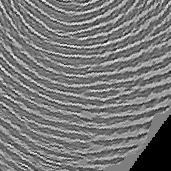}
	}         
	\caption{Examples of training fingerprint patches.}
	\label{fig:train_eg}
	\vspace*{-5pt}
\end{figure}

\begin{figure}[t]
	\centering
		\subfigure[]{
		\includegraphics[clip, trim=0cm 0cm 0cm 0cm, width=0.21\linewidth]{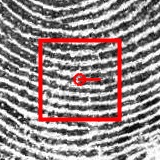}
	}  
	\subfigure[][]{
		\includegraphics[clip, trim=0cm 0cm 0cm 0cm, width=0.21\linewidth]{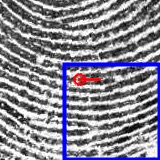}
	}
	\subfigure[]{
		\includegraphics[clip, trim=0cm 0cm 0cm 0cm, width=0.21\linewidth]{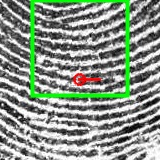}
	}     
	\subfigure[]{
		\includegraphics[clip, trim=0cm 0cm 0cm 0cm, width=0.21\linewidth]{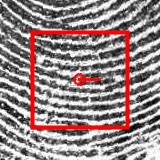}
	}       
	\caption{Four different patch types used in \cite{Cao2018PAMI} for descriptor extraction. Patch types in (a)-(c) were determined to be the best combination in terms of identification accuracy. The patches in (a), (b) and (d) are of sizes $96\times 96$ pixels while the patch in (c) is of size $80\times 80$ pixels. In this paper, we used patch types in (b), (c) and (d) for descriptor extraction because all of them are of the same size and hence no resizing is needed.}
	\label{fig:patch_types}
	\vspace*{-5pt}
\end{figure}

A minutia descriptor contains attributes of the minutia
based on the image characteristics in its neighborhood.
Salient descriptors are needed to establish 
 minutiae correspondences and compute the similarity between a latent and
reference prints. Instead of specifying the descriptor in an ad hoc manner, Cao and Jain \cite{Cao2018PAMI} trained ConvNets to learn
the descriptor from local fingerprint patches around a minutia and showed its performance. In this paper, we improve both the distinctiveness of the descriptor and efficiency of descriptor extraction. 

Cao and Jain \cite{Cao2018PAMI} extracted around 800K  $160 \times 160$ fingerprint patches from around 50K minutiae from images in a large fingerprint longitudinal database  \cite{Yoon2015PNAS}, to train the ConvNets. On average, there are  around 16 fingerprint patches around each minutia for training. Cao and Jain also reported that descriptors extracted from enhanced latent images give better performance. In order to augment the training dataset and improve the descriptor distinctiveness between enhanced latent and original rolled prints, we use patches from  both the original and enhanced fingerprint patches for training ConvNets; this results in around 1.6 million fingerprint patches for training. Fig. \ref{fig:train_eg} shows some example patches from the training dataset;  Figs. \ref{fig:train_eg} (a) to (d) are from the original image and (e) to (h) are from the corresponding enhanced images.

Location and size of the patches were also evaluated in \cite{Cao2018PAMI}. The three patch types shown in Figs. \ref{fig:patch_types} (a)-(c) were determined to be the best patch types via forward sequential selection. The two patches in Figs. \ref{fig:patch_types} (b) and (c) are of the same size ($96\times 96$ pixels) while the patch in Fig. \ref{fig:patch_types} (a) is of size $80 \times 80$ pixels. All the patches had to be resized to  $160\times 160$ pixels in \cite{Cao2018PAMI}, by bilinear interpolation. In order to avoid resizing, we train a ConvNet on $96\times 96$  images and use patch types in Figs. \ref{fig:patch_types} (b)-(d) for descriptor extraction. Note that the patch in Fig. \ref{fig:patch_types} (d) is minutiae centered whereas patches in Figs. \ref{fig:patch_types} (b) and (c) are offset from the center minutiae.
 
Among the various ConvNet architectures  \cite{VGG},  \cite{GoogLeNet},  \cite{ResNet},  \cite{Mobilenet}, MobileNet-v1 \cite{Mobilenet} uses depth-wise separable convolutions,  resulting in a 
drastic reduction in model size and training/evaluation times while providing good recognition performance. The number of original model parameters to be trained in MobileNet-v1 (4.24M), is significantly
smaller than the number of model parameters in Inception-v3 (23.2M) and VGG (138M), requiring significantly lower
efforts in terms of regularization and data augmentation, to prevent overfitting. For these reasons, we utilize the MobileNet-v1 architecture  due to its fast inference speed.
In order to feed $96 \times 96$ fingerprint patches for training and to reduce the descriptor length, we make the following modifications  to the MobileNet-v1 architecture: i) change the input image size to $96 \times 96 \times 1$, ii) remove the last two convolutional layers to accommodate the smaller input sizes, and iii) add a fully connected layer, also called the embedding layer, before the classifier layer to obtain a compact feature representation. The modified architecture is shown in Table \ref{tab:MobileNet}.

For each of the three patch types shown in Figs. \ref{fig:patch_types} (b)-(d), 1.6 million fingerprint patches are used to train a MobileNet. 
Given a  fingerprint patch around a virtual minutiae, the output ($l$-dimensional feature vector) of the last fully connected layer  is considered as the virtual minutiae descriptor. In the experiments, three value of $l$, namely $l=32, 64$, and $128$, are investigated. The concatenation of the three outputs for the same virtual minutiae is regarded as the descriptor with length $l_d$ = $3\times l$.  
The set of virtual minutiae and the associated descriptors define a texture template.

\subsection{Texture Template Matching}

\begin{figure}[t]
    \centering
        \includegraphics[clip, trim=0cm 0cm 0cm 0cm, width=0.5\linewidth]{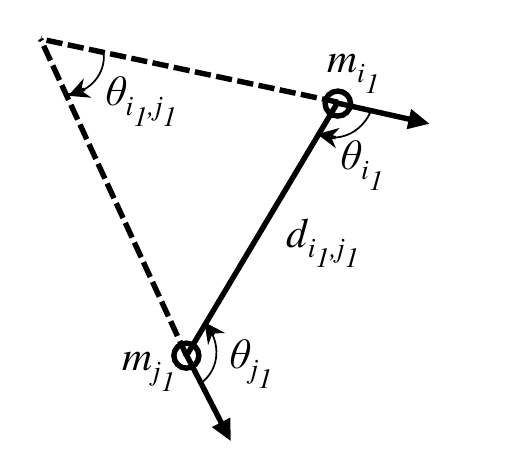}
    \caption{Illustration of second-order graph matching \cite{Cao2018PAMI}, where $m_{i_1}$ and $m_{j_1}$ are the two virtual minutiae, $d_{i_1,j_1}$ is the Euclidean distance between $m_{i_1}$ and $m_{j_1}$,  and $\theta_{i_1,j_1}$, $\theta_{i_1}$ and $\theta_{j_1}$ are the three angles formed by two virtual minutiae orientations and the line segment connecting  $m_{i_1}$ and $m_{j_1}$. }
      \label{fig:second-order}
      \setlength{\belowdisplayskip}{-6pt}
\setlength{\belowdisplayshortskip}{-6pt}
\vspace*{-5pt}
\end{figure}

\begin{table}[tp]
\caption{MobileNet Architecture, where $l$ is the length of feature vector output by the MobileNet and $c$ is the number of classes used for training.}
\begin{center}
\begin{tabular}{|c|c|c|}
\hline
Type / Stride  & Filter Shape & Input Size\\
\hline
Conv / s2 & 3$\times$3$\times$1$\times$32  & 96$\times$96$\times$1 \\
\hline
Conv dw / s1 & 3$\times$3$\times$32 dw &48$\times$48$\times$32 \\
\hline
Conv / s1 &1$\times$1$\times$32$\times$64 &48$\times$48$\times$32 \\
\hline
Conv dw / s2 & 3$\times$3$\times$64 dw &48$\times$48$\times$64 \\
\hline
Conv / s1 &1$\times$1$\times$64$\times$128 &24$\times$24$\times$64 \\
\hline
Conv dw / s1 &3$\times$3$\times$128 dw &56$\times$24$\times$128 \\
\hline
Conv / s1 &1$\times$1$\times$128$\times$128 &24$\times$24$\times$128  \\
\hline
Conv dw / s2 &3$\times$3$\times$128 dw &24$\times$24$\times$128 \\
\hline
Conv / s1 &1$\times$1$\times$128$\times$256 &12$\times$12$\times$128 \\
\hline
Conv dw / s1 &3$\times$3$\times$256 dw &12$\times$12$\times$256 \\
\hline
Conv / s1 &1$\times$1$\times$256$\times$256 &12$\times$12$\times$256 \\
\hline
Conv dw / s2 &3$\times$3$\times$256 dw &12$\times$12$\times$256 \\
\hline
Conv / s1 & 1$\times$1$\times$256$\times$512 &6$\times$6$\times$256 \\
\hline
5$\times $Conv dw / s1& 3$\times$3$\times$512 dw &6$\times$6$\times$512 \\ 
\hline
Conv / s1 &1$\times$1$\times$512$\times$512 &6$\times$6$\times$512 \\
\hline
Conv dw / s2 &3$\times$3$\times$512 dw &6$\times$6$\times$512 \\
\hline
Avg Pool / s1 & Pool 6$\times$6  & 6$\times$6$\times$512\\
\hline
FC / s1  & 512$\times l$ &  1$\times 1 \times 512$ \\
\hline
FC / s1  &  $l \times c $&  1$\times 1 \times l$ \\
\hline
Softmax / s1 & Classifier & 1 $\times $ 1 $\times  c$\\
\hline
\end{tabular}
\end{center}
\label{tab:MobileNet}
\vspace*{-5pt}
\end{table}%

\label{sec:matching}
The algorithm for comparing two texture templates, one from latent and the other from a reference print, as proposed in \cite{Cao2018PAMI}  can be summarized as: i) compute pair-wise similarities between the latent and reference print virtual minutiae descriptors using cosine similarity;  normalize the similarity matrix, ii) select the top $N$ ($N=200$) virtual minutiae correspondences based on the normalized similarity matrix, iii) remove false minutiae correspondences using second-order graph matching, iv) further remove false minutiae correspondences using third-order graph matching, and (v) finally compute the overall similarity between the two texture templates  based on final minutiae correspondences. Although the second-order graph matching can remove most false correspondences, it is still time-consuming as it involves $N(N-1)/2$ computations. Furthermore, since the locations of virtual minutiae do not have large variations due to the raster scan, the larger complexity third order graph matching does not help too much in virtual minutiae correspondences.

\begin{figure}[t]
	\centering
	\subfigure[]{
		\includegraphics[clip, trim=2cm 1cm 1cm 1cm, width=0.85\linewidth]{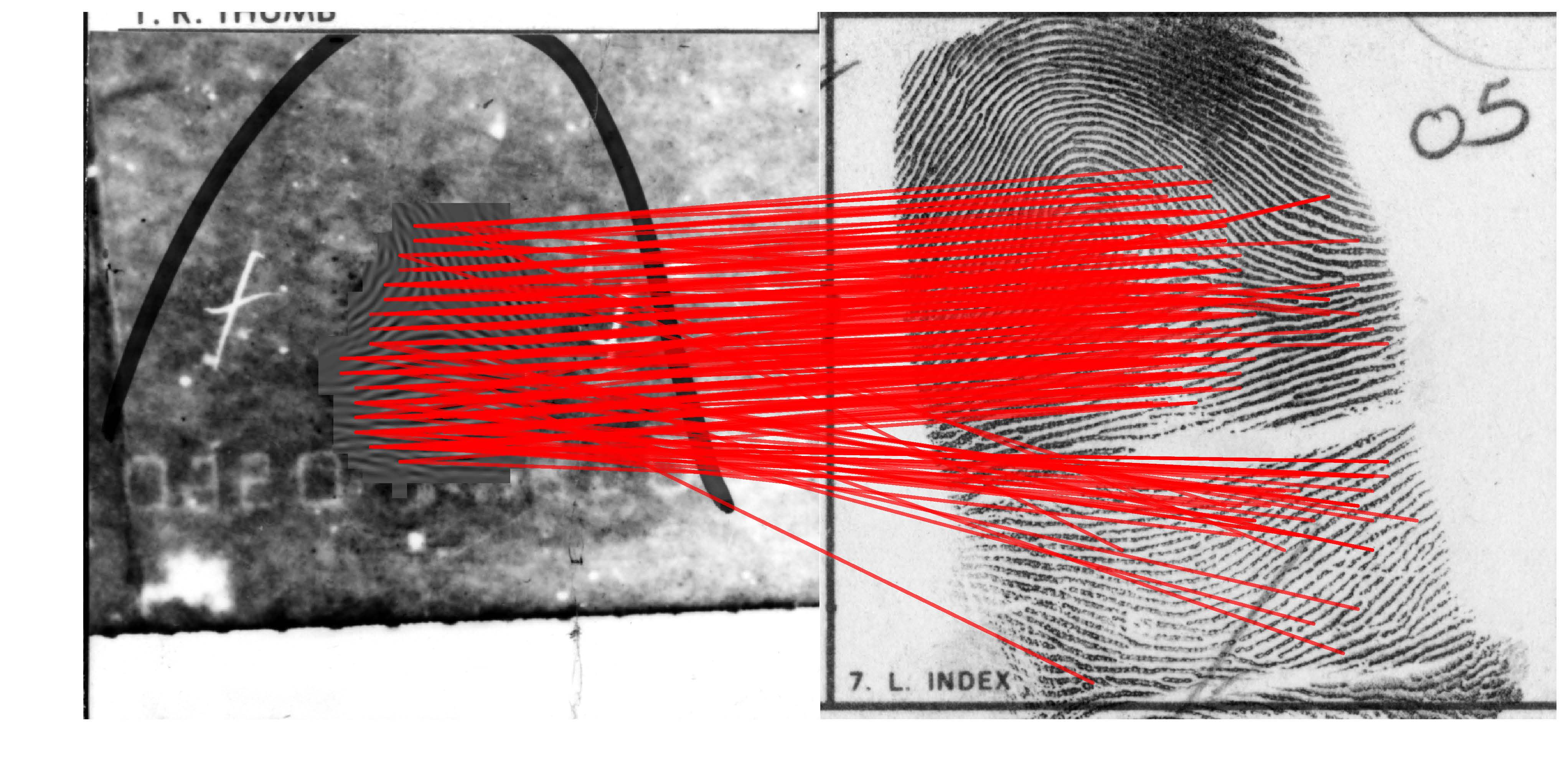}
	}
	\subfigure[]{
		\includegraphics[clip, trim=2cm 1cm 1cm 1cm, width=0.85\linewidth]{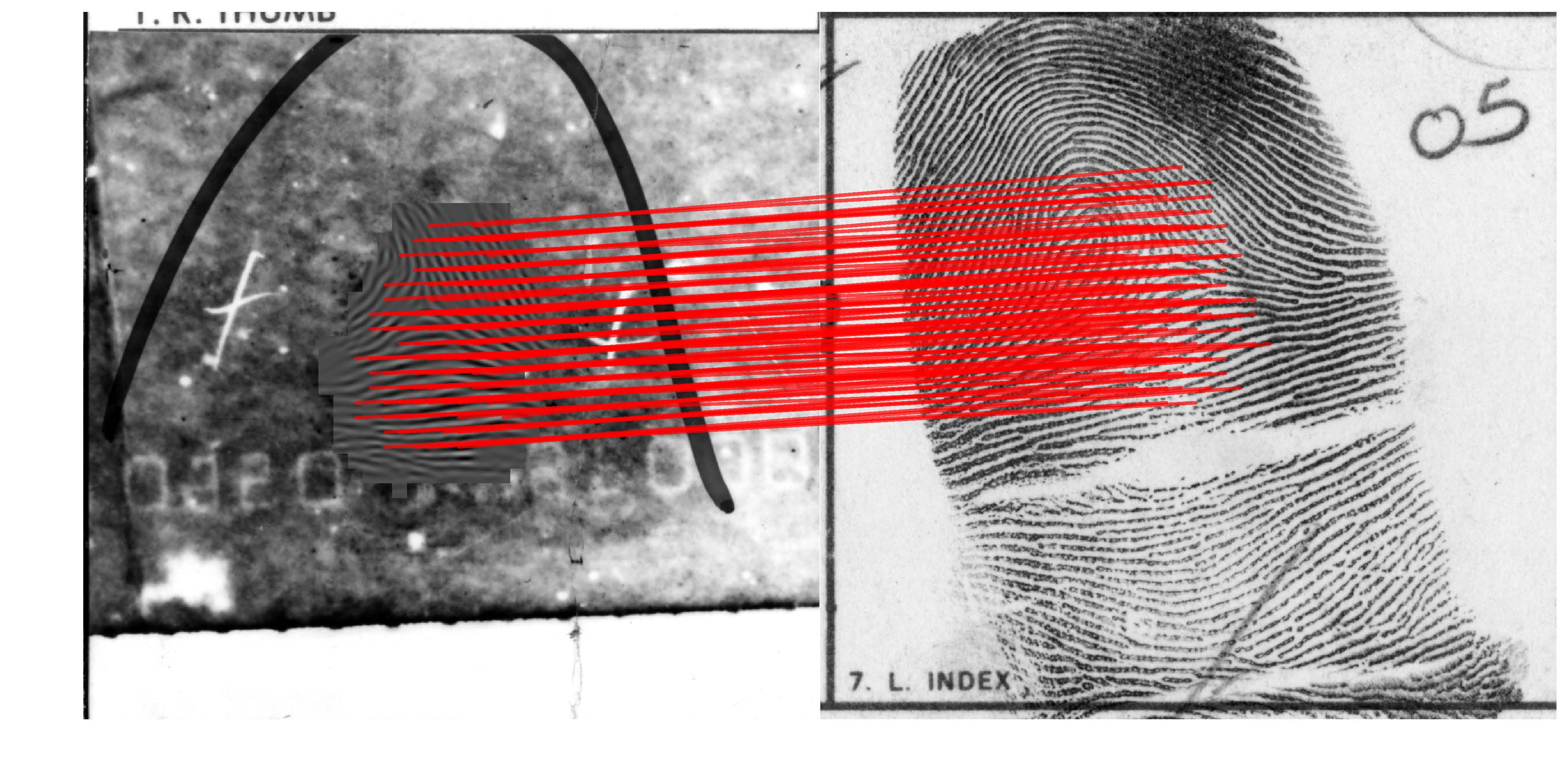}
	}   
	\subfigure[]{
		\includegraphics[clip, trim=2cm 1cm 1cm 1cm, width=0.85\linewidth]{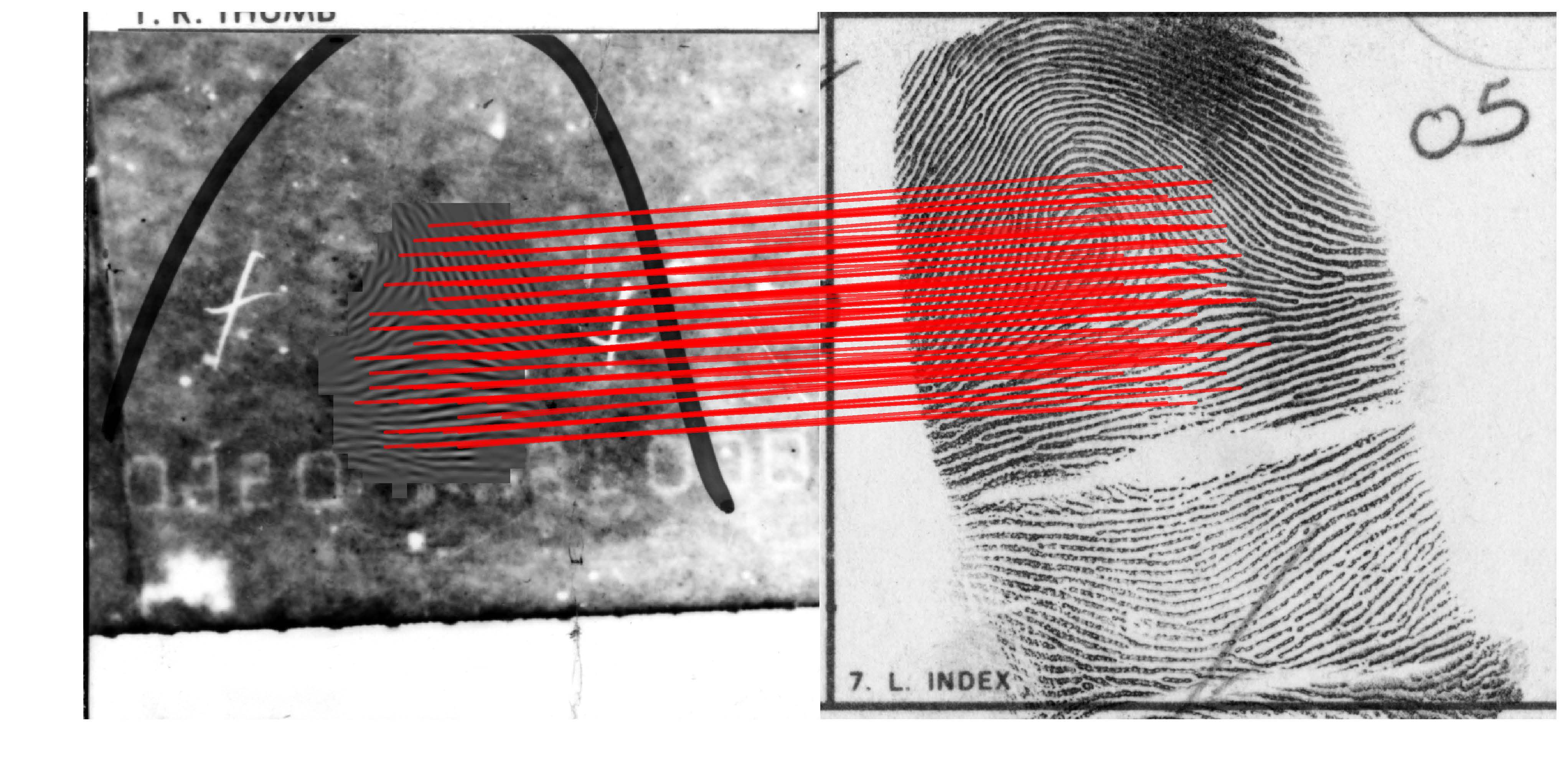}
	}                  
	\caption{Illustration of virtual minutiae correspondences by the proposed graph matching strategy. Figures in  (a), (b) and (c) show the top $N=200$ minutiae correspondences,   88 virtual minutiae correspondences by using the modified second order graph matching and 73 virtual minutiae correspondences by applying the second order graph matching used in \cite{Cao2018PAMI} to the 88 minutiae correspondences in (b).}
	\label{fig:correspondences}
	\vspace*{-5pt}
\end{figure}

As shown in Fig. \ref{fig:second-order}, the computation of the compatibility between two virtual minutiae pairs in second-order graph matching involves one Euclidean distance computation and three angular distance computations. 
In our preliminary experiments, we found that the Euclidean distance alone is good enough to remove most false correspondences as shown in Fig. \ref{fig:correspondences}.
The approach we propose here is to use a modified second-order graph matching to remove most false virtual minutiae correspondences and use the second-order graph matching in \cite{Cao2018PAMI} to get the final virtual minutiae correspondences

 An overview of the modified virtual minutiae matching algorithm is illustrated in $\mathbf{Algorithm       \ \ref{alg:search}}$.  The details of the modified second-order graph matching are as following.  Suppose  $\{i=(i_1,i_2)\}_{i=1}^{N}$ is the set of $N$ selected minutiae correspondences between a latent $L$ and a rolled print $R$, where $i_1$ and $i_2$ denote the $i^{th}$ correspondence between the $i_1^{th}$ and $i_2^{th}$ virtual minutiae in the latent $F_l$ and the rolled print $F_r$, respectively.  Given two  minutiae correspondences $(i_1,i_2)$ and $(j_1,j_2)$, $H_{i,j}^2$ ($H^2 \in R^{N\times N}$) measures the compatibility between $(i_1,j_1)$ from the latent and $(i_2,j_2)$ from the rolled prints:
\begin{align}
H^2_{i,j} &= Z(D_{i,j},\mu,\tau,t),
%s.t. \  &\forall i, \sum\nolimits_{i}X_{i,j}\leqslant 1 \nonumber \\
% &\forall j,  \sum\nolimits_{j}X_{i,j}\leqslant 1 \nonumber
\end{align}
where $D_{i,j} = |d_{i_1,j_1} - d_{i_2,j_2}|$ and $Z$ is a truncated sigmoid function:

\begin{equation}
Z(v,\mu,\tau, t) =	\left.
\begin{cases}
 \frac{1}{1+e^{-\tau(v-\mu)}}, & \text{   if } v \leq t, \\
 0, & \text{   otherwise. }
\end{cases}\right.
\end{equation}
Here $\mu,\tau$ and $t$ are the parameters of function $Z$.

 The goal of graph matching is to find an $N$-dimensional correspondence vector $Y$, where the $i^{th}$ element ($Y_i$) indicates whether $i_1$  is assigned to $i_2$  ($Y_i = 1$) or not ($Y_i = 0$). This can be represented in terms of  maximizing the following objective function:
 \begin{equation}
\label{eq:objective_simple_two}
S_2(Y) = \sum_{i,j}  \\ H^2_{i,j} Y_i Y_j.
\end{equation}
A strategy of power
iteration followed by discretization used in \cite{Cao2018PAMI} is used to remove false minutiae correspondences.

Suppose  $\{i=(i_1,i_2)\}_{i=1}^{n}$  represent the final $n$ matched minutiae correspondences between $F_l$ and $F_r$. The similarity $S$  between $F_l$ and $F_r$ is defined as:
\begin{equation}
\label{eq:minutiae_similarity}
S = \sum_{i=1}^nDesSim(i_1,i_2),
\end{equation}
where $DesSim(i_1,i_2)$ is the descriptor similarity between $i_1^{th}$ virtual minutiae in latent template $F_l$ and $i_2^{th}$ virtual minutiae in reference template  $F_r$.

\begin{algorithm}
	\caption{Modified virtual minutiae matching algorithm}\label{alg:search}
	\begin{algorithmic}[1]
		%\Procedure{}{}
		\State \textbf{Input:} Latent template $F_l$ and reference template $F_r$ 
		\State \textbf{Output:} Similarity between $F_l$ and $F_r$ 
		\State 	Compute descriptor similarity matrix\;
		\State	Normalize similarity matrix\;
		\State  Select the top $N$ minutiae correspondences based on the normalized similarity matrix\;
		\State 	Construct $H^2$ based on these $N$ minutiae correspondences\;
		\State	Remove false correspondences using modified second-order graph matching\;
		\State  Further remove false correspondences using original second-order graph matching\;
		\State  Compute similarity between $F_l$ and $F_r$\;
		%	\EndProcedure
	\end{algorithmic}
\end{algorithm}

%\begin{algorithm}
%	\caption{Modified virtual minutiae matching algorithm}\label{alg:search}
%	\begin{algorithmic}[1]
%		%\Procedure{}{}
%		\State \textbf{Input:} Latent template $T^l$ and $M$ reference templates $\{T^r_j\}_{j=1}^M$ 
%		\State \textbf{Output:} Top $n$ candidates and their similarities
%		\For{ each $T^r_j$}
%		\State 	Compute descriptor similarity matrix\;
%		\State	Normalize similarity matrix;
%		\State       Select the top $N$ minutiae correspondences based on Normalized similarity matrix\;
%		\State 	Construct $H^2$ based on these $N$ minutiae correspondences\;
%		\State	Remove false correspondences using modified second-order graph matching\;
%		\State       Compute similarity between $T^l$ and $T^r_j$\;
%		\EndFor 
%		
%		\State	Retrieve top $K$ candidates based on their similarities\;
%		\State	Refine candidate order using third-order graph matching\;
%		\State 	Output final top $n$ candidates and their similarities.
%	%	\EndProcedure
%	\end{algorithmic}
%\end{algorithm}

 \begin{figure}[t] 
	\centering
	\subfigure[]{
		\includegraphics[clip, trim=6cm 8cm 7cm 8cm, width=0.35\linewidth]{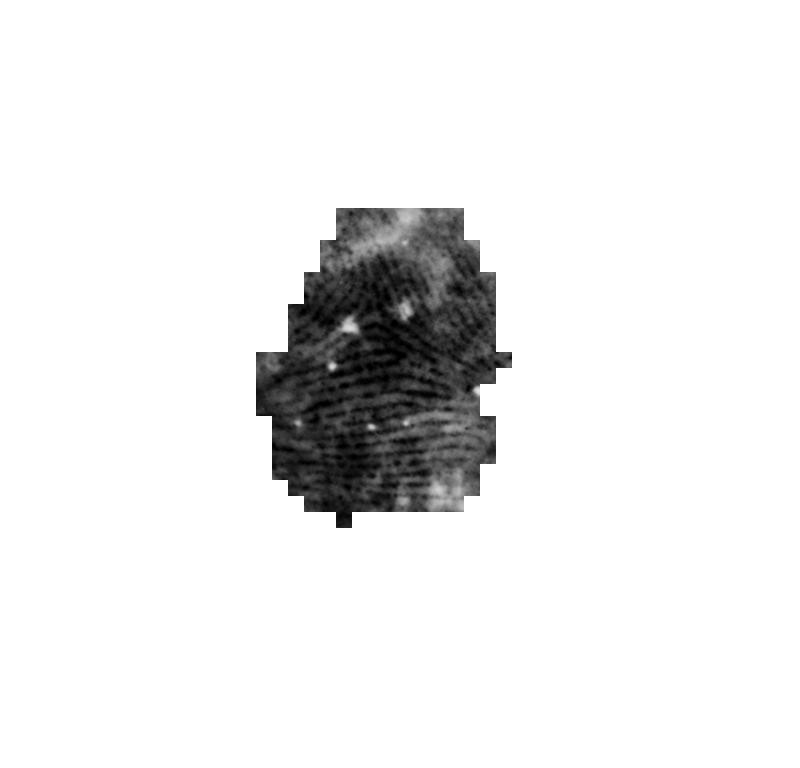}
	} \hspace{1 cm}
	\subfigure[]{
		\includegraphics[clip, trim=6cm 8cm 7cm 8cm, width=0.35\linewidth]{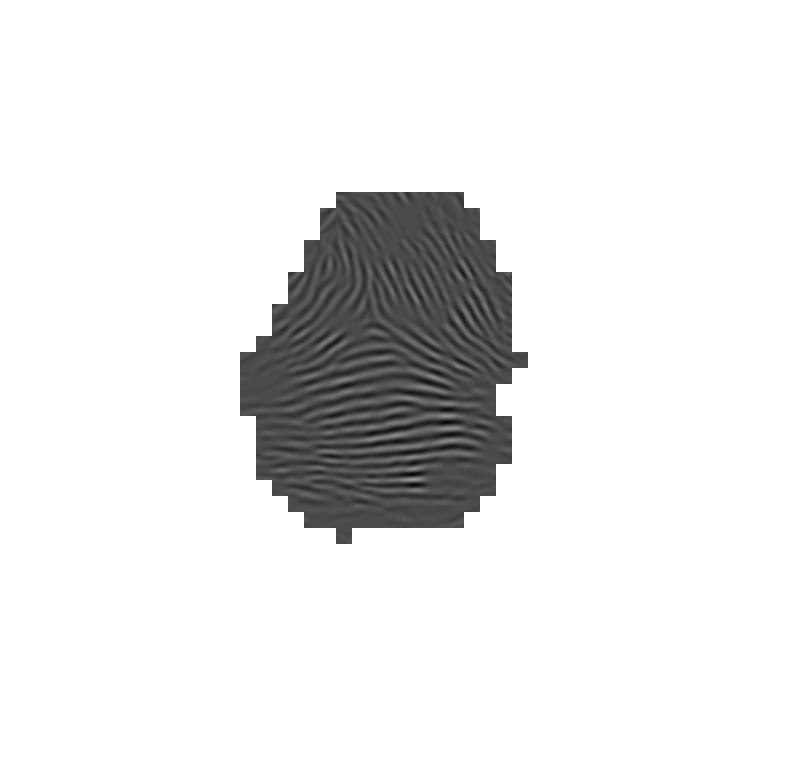}
	}     \hspace{1 cm}
	\subfigure[]{
		\includegraphics[clip, trim=0cm 0cm 0cm 0cm, width=0.22\linewidth]{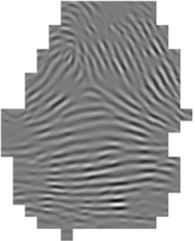}
	}   \hspace{1.5 cm}
	\subfigure[]{
		\includegraphics[clip, trim=6cm 8cm 7cm 8cm, width=0.35\linewidth]{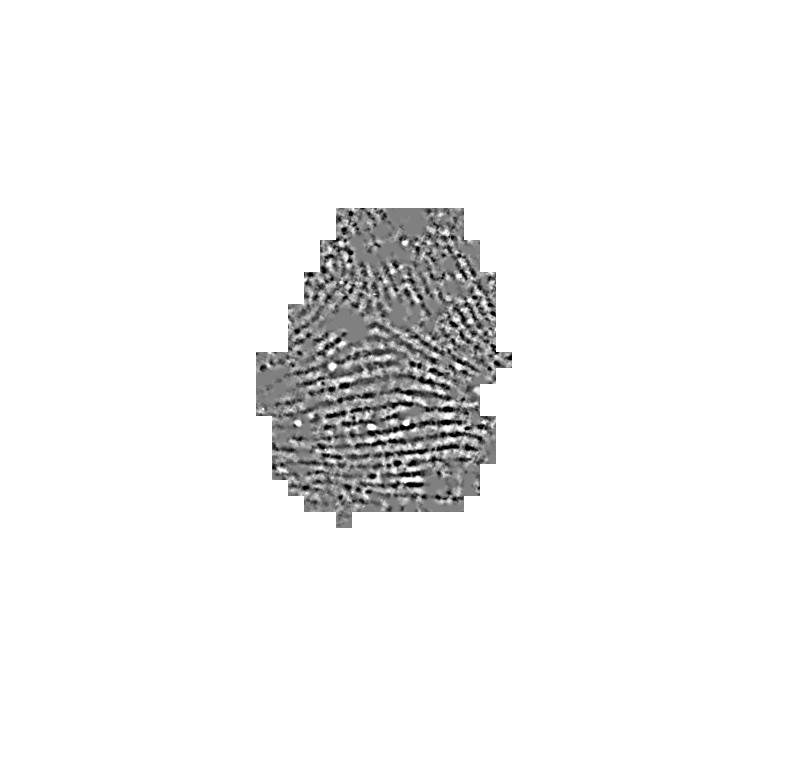}
	}           
	\caption{Illustration of processing strategies applied to the input latent shown in (a) for virtual minutiae descriptor extraction.  (b) and (c) are the two enhanced latent fingerprint images used for minutiae template 1 and 2 extraction in \cite{Cao2018PAMI}, and (d) is the texture component after decomposition \cite{Cao2014PAMI}. }
	\label{fig:input_images}
	\vspace*{-5pt}
\end{figure}

\subsection{Constructing texture templates}
\label{sec:templates}
The locations and orientations of virtual minutiae are determined by the ROI and ridge flow, while the descriptors depend on the images input to the trained ConvNets.  Three different processed latent images for each latent are investigated for texture template construction. Two different enhancement algorithms were proposed in \cite{Cao2018PAMI} and the resulting two minutiae sets were extracted, one per enhanced images.  Both of the enhanced images for virtual minutiae descriptor extraction are also investigated here. However, the fingerprint enhancement performance critically depends  on the estimates of  ridge flow and ridge spacing. If ridge flow or ridge spacing are not estimated correctly, spurious ridge structures are created in the enhanced images.  In addition to the two enhanced images, we also consider the texture image obtained by image decomposition \cite{Cao2014PAMI}, which essentially removes large scale background noise and enhances ridge contrast.  Figs. \ref{fig:input_images} (b)-(d) illustrate the three processed images for the input latent image in Fig. \ref{fig:input_images} (a). For each latent, three different texture templates, $T_{e_1}$,  $T_{e_2}$ and $T_{t}$, can then be extracted. Given a latent to reference print pair, three texture template similarities are computed  which are then fused to improve the overall latent recognition performance.

\section{Experimental Results}
The proposed texture template matching algorithm is evaluated on  NIST SD27 latent database, which consists of 88 good quality, 85 bad quality and 85 ugly quality latent fingerprint images. For latent search experiments,  10,000 reference prints\footnote{Results for the larger gallery of 100K reference prints are not yet available at the time of submitting this paper.}, including the 258 mates of NIST SD27 and others from NIST SD14, are used as the gallery.

\subsection{Descriptor length evaluation}
\begin{figure}[t] 
	\centering
	\subfigure[]{
		\includegraphics[clip, trim=5cm 2cm 4cm 2cm, width=0.45\linewidth]{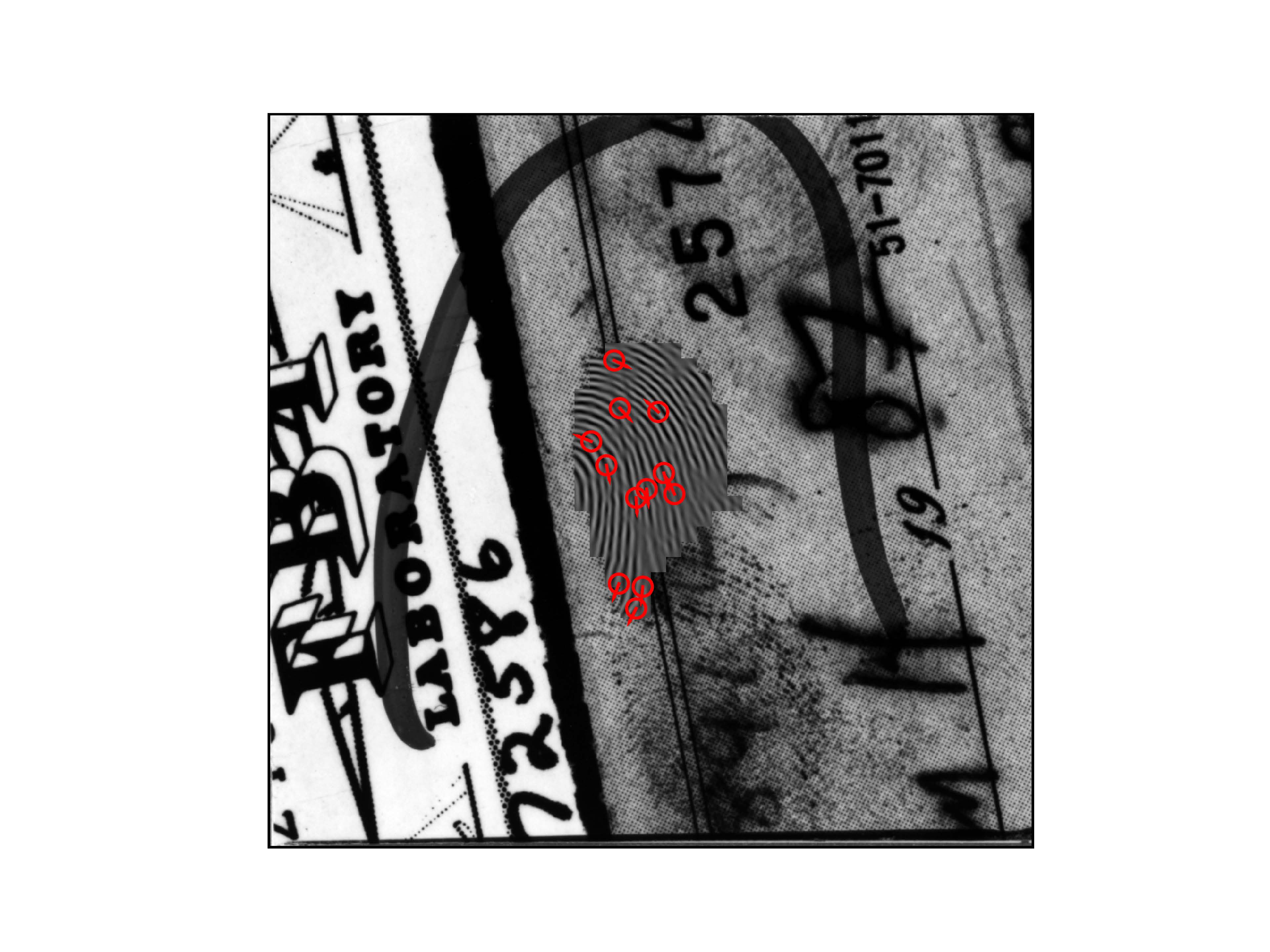}
	}
	\subfigure[]{
		\includegraphics[clip, trim=5cm 2cm 4cm 2cm, width=0.45\linewidth]{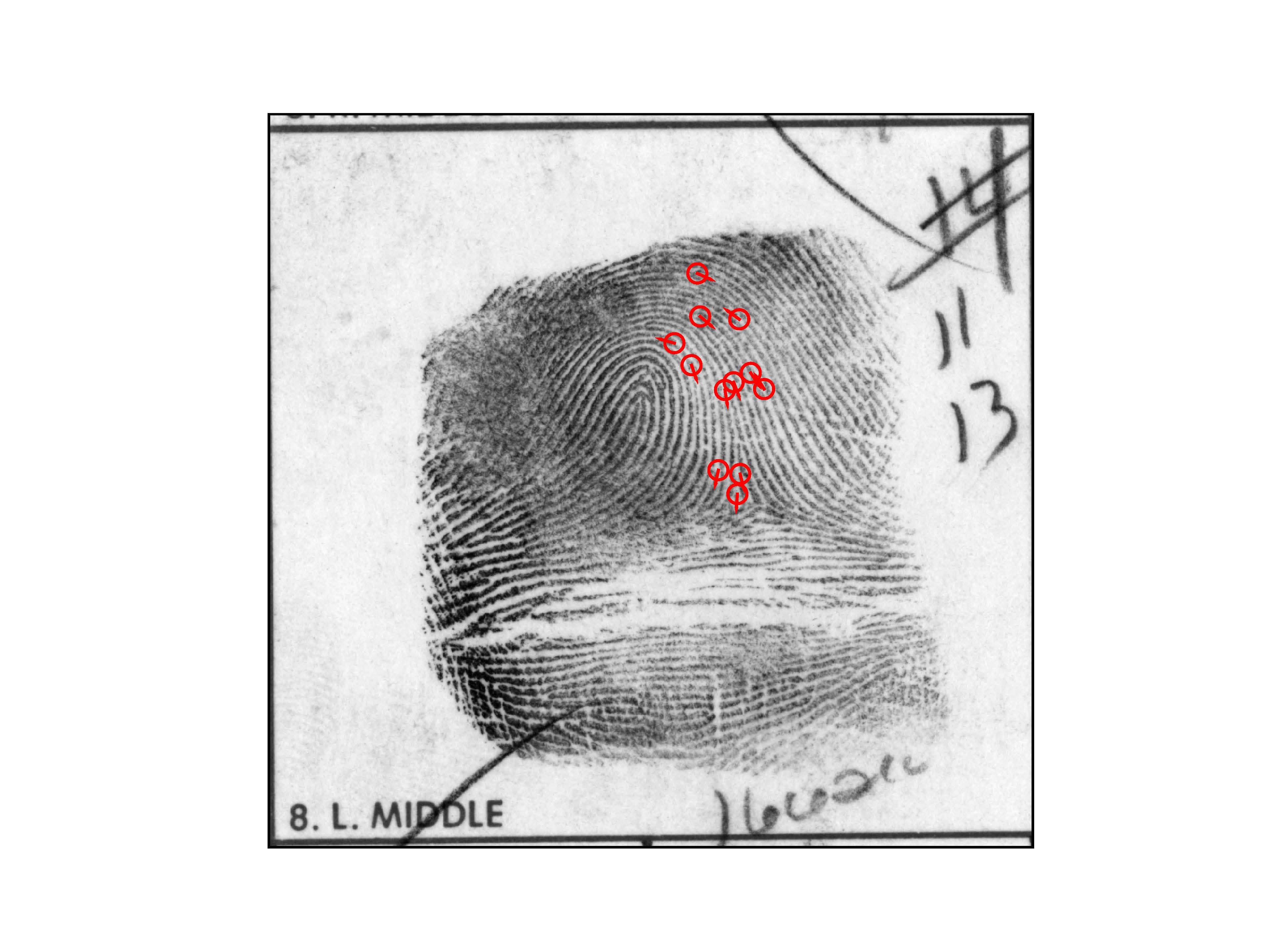}
	}             
	\caption{Manually marked minutiae correspondences are used to evaluate the proposed virtual minutiae descriptors. (a) Manually marked latent minutiae shown on the enhanced latent image and (b)  manually marked rolled minutiae shown on mated rolled print image.}
	\label{fig:manual_minutiae}
	\vspace*{-5pt}
\end{figure}

\begin{figure}[t]
    \centering
        \includegraphics[clip, trim=0cm 0cm 0cm 0cm, width=0.9\linewidth]{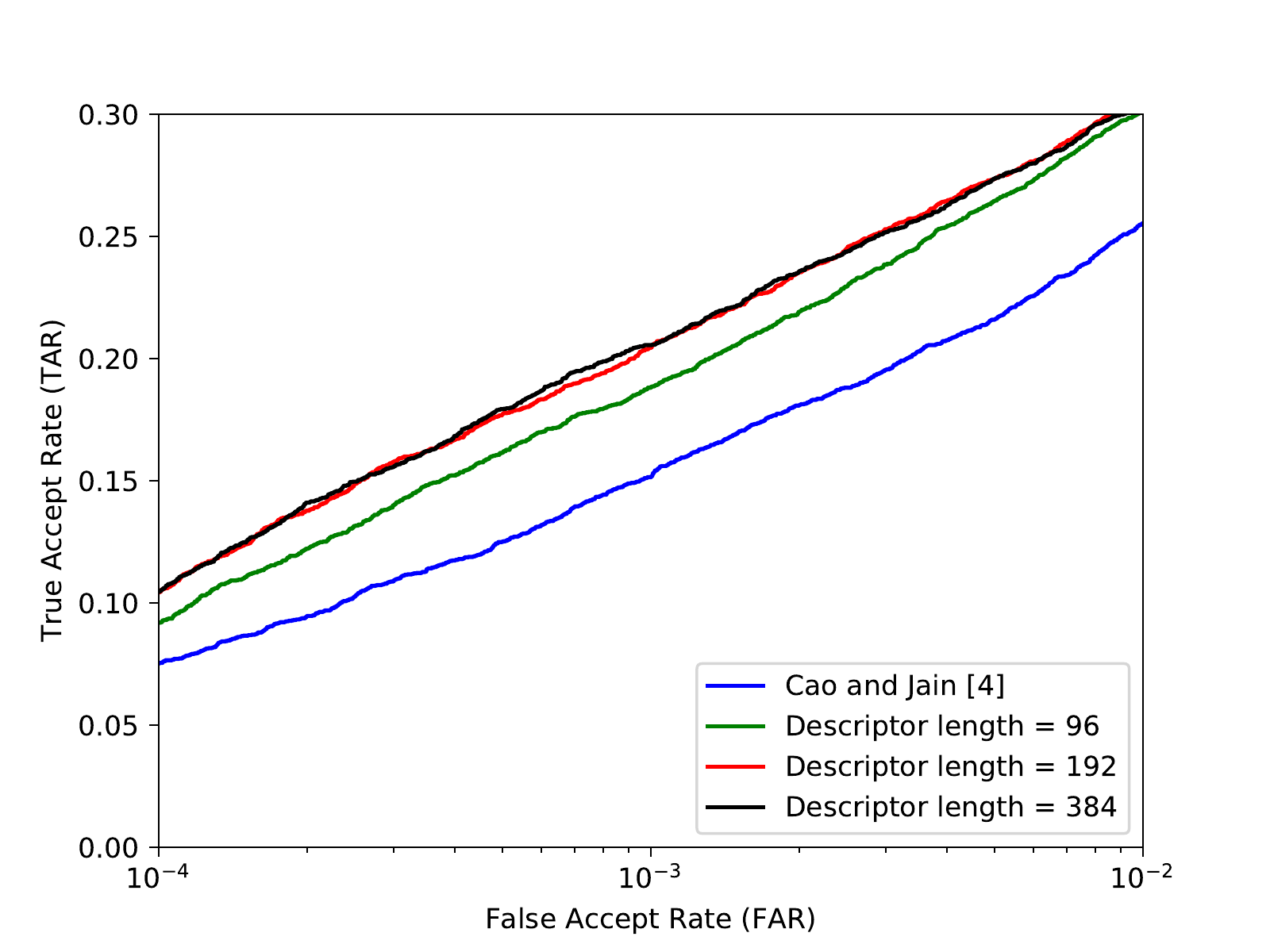}
    \caption{Receiver Operating Characteristic (ROC) curves under different descriptor lengths.}
      \label{fig:verification}
\vspace*{-5pt}
\end{figure}

The performance of the new virtual minutiae descriptors with different feature lengths, namely, $l_d$ = 96 ($l=32$), 192 ($l=64$) and 384 ($l=128$), are evaluated by verification performance based on manually marked minutiae correspondences on NIST SD27\footnote{NIST SD27 dataset  is no longer available for  download from the NIST site.} \cite{NISTDB27}. A total of 5,460 minutiae correspondences between the latent images and the mated rolled fingerprint images were provided with the NIST SD27 database.  The average numbers of manually marked minutiae correspondences on 88 good quality, 85 bad quality and 85 ugly quality latent images are 31, 18 and 14, respectively.  For a fair comparison with the descriptors used in \cite{Cao2018PAMI}, the descriptors for the latents are extracted on the same enhanced images as in \cite{Cao2018PAMI} while the descriptors of the mated rolled prints are extracted on the original rolled prints.  Fig. \ref{fig:manual_minutiae} shows the manually marked minutiae on an enhanced latent image and its mated rolled prints. The genuine scores are computed using the similarities of descriptors from the manually marked minutiae correspondences while the impostor scores are computed using the similarities of descriptors from the different minutiae. Thus, a total of 5,460 genuine scores and around 30 million impostor scores ($5,460\times 5,460$) are computed.  Fig. \ref{fig:verification} compares the Receiver Operating Characteristic (ROC) curves of different descriptor lengths as well as the descriptors from \cite{Cao2018PAMI} which used a descriptor length of 384. Note that the ROC curves of descriptors with feature lengths 384 and 192 are very close to each other, slightly better than descriptors with feature length 96 and significantly better than the descriptors in \cite{Cao2018PAMI}.  This can be explained by the use of both enhanced and original fingerprint patches along with a more appropriate ConvNet architecture for training.

 \begin{figure*}[t] 
	\centering
	\subfigure[]{
		\includegraphics[clip, trim=4cm 8.5cm 4cm 9cm, width=0.35\linewidth]{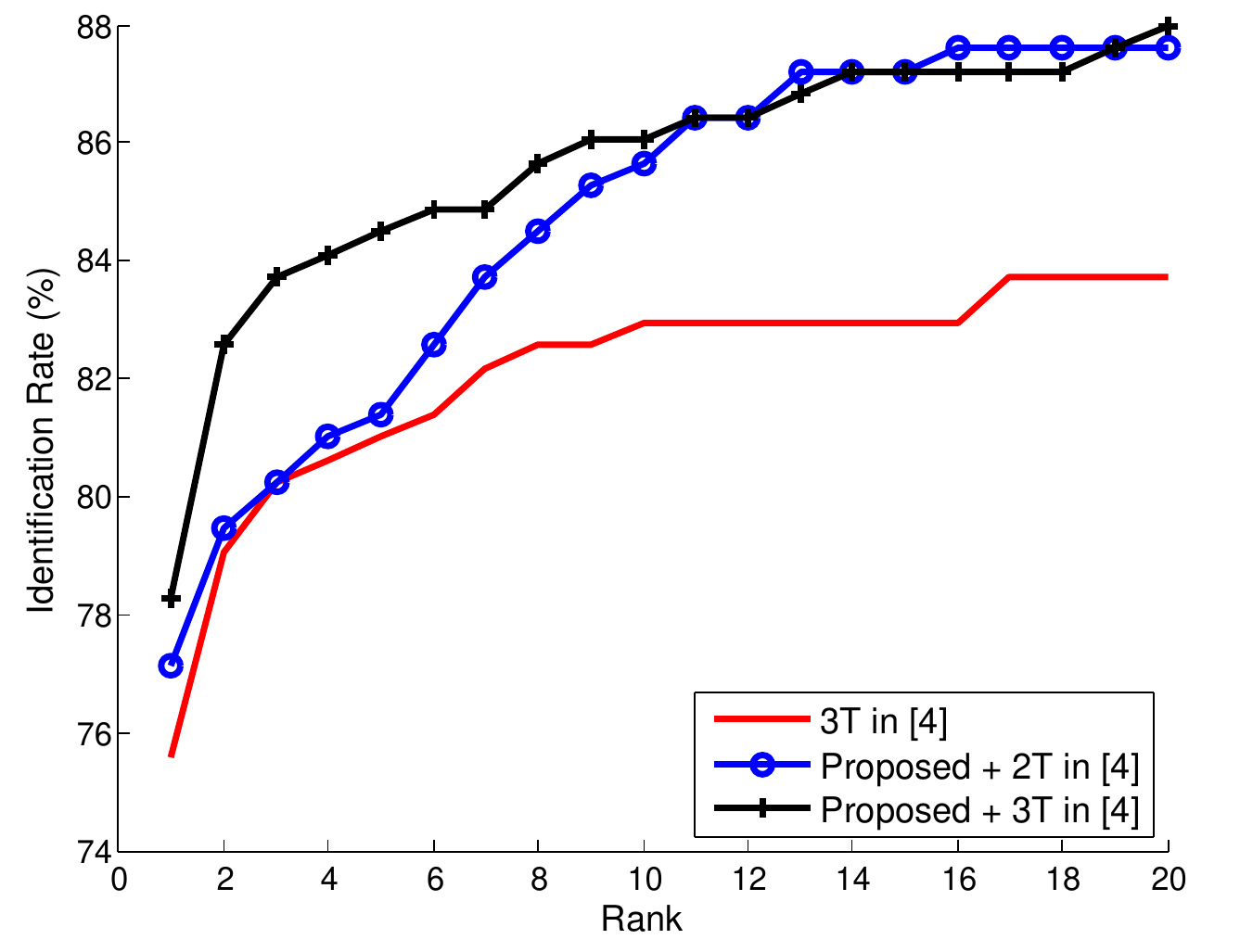}
	} \hspace{1cm}
	\subfigure[]{
		\includegraphics[clip, trim=4cm 8.5cm 4cm 9cm, width=0.35\linewidth]{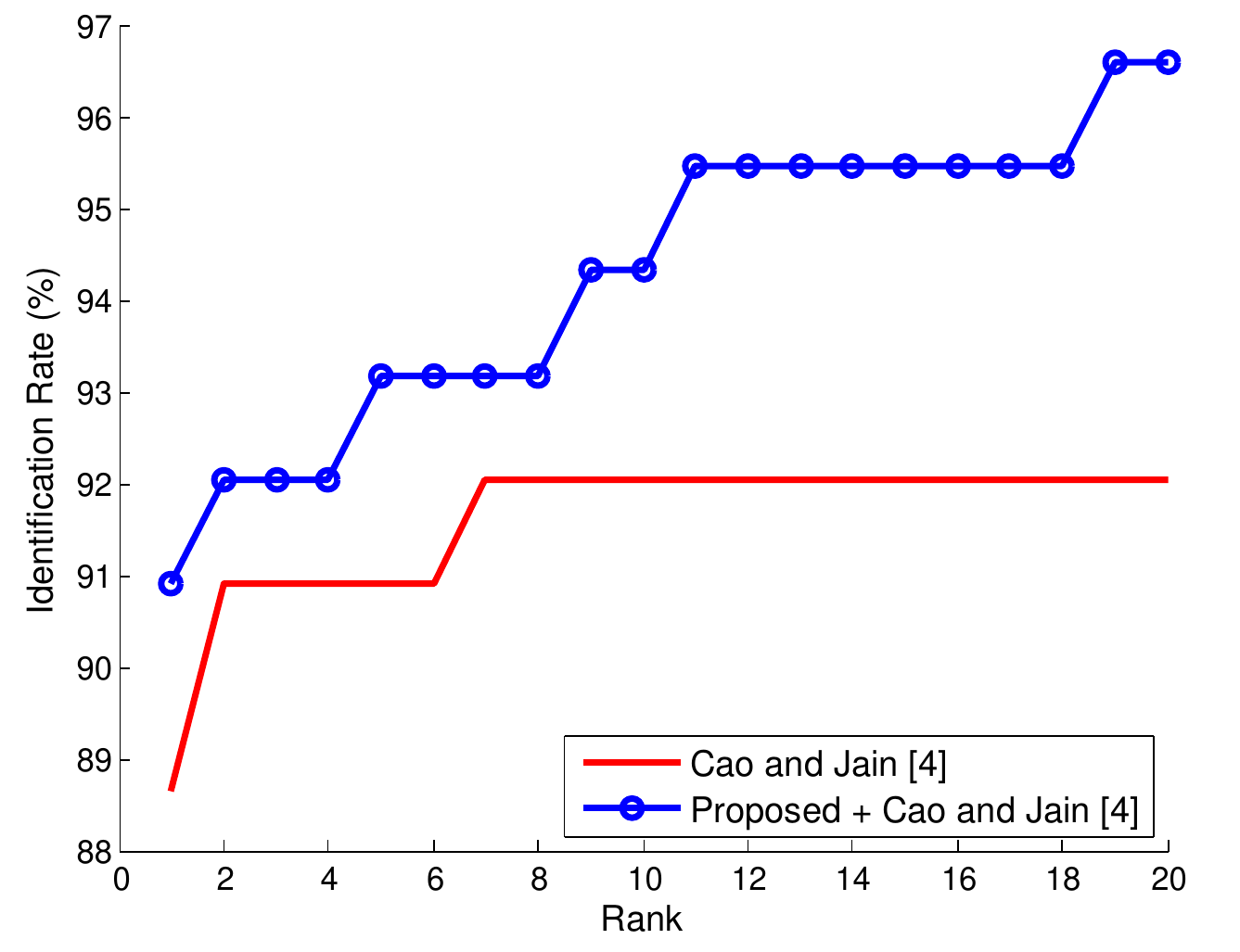}
	}     \hspace{1cm}
	\subfigure[]{
		\includegraphics[clip, trim=4cm 8.5cm 4cm 9cm, width=0.35\linewidth]{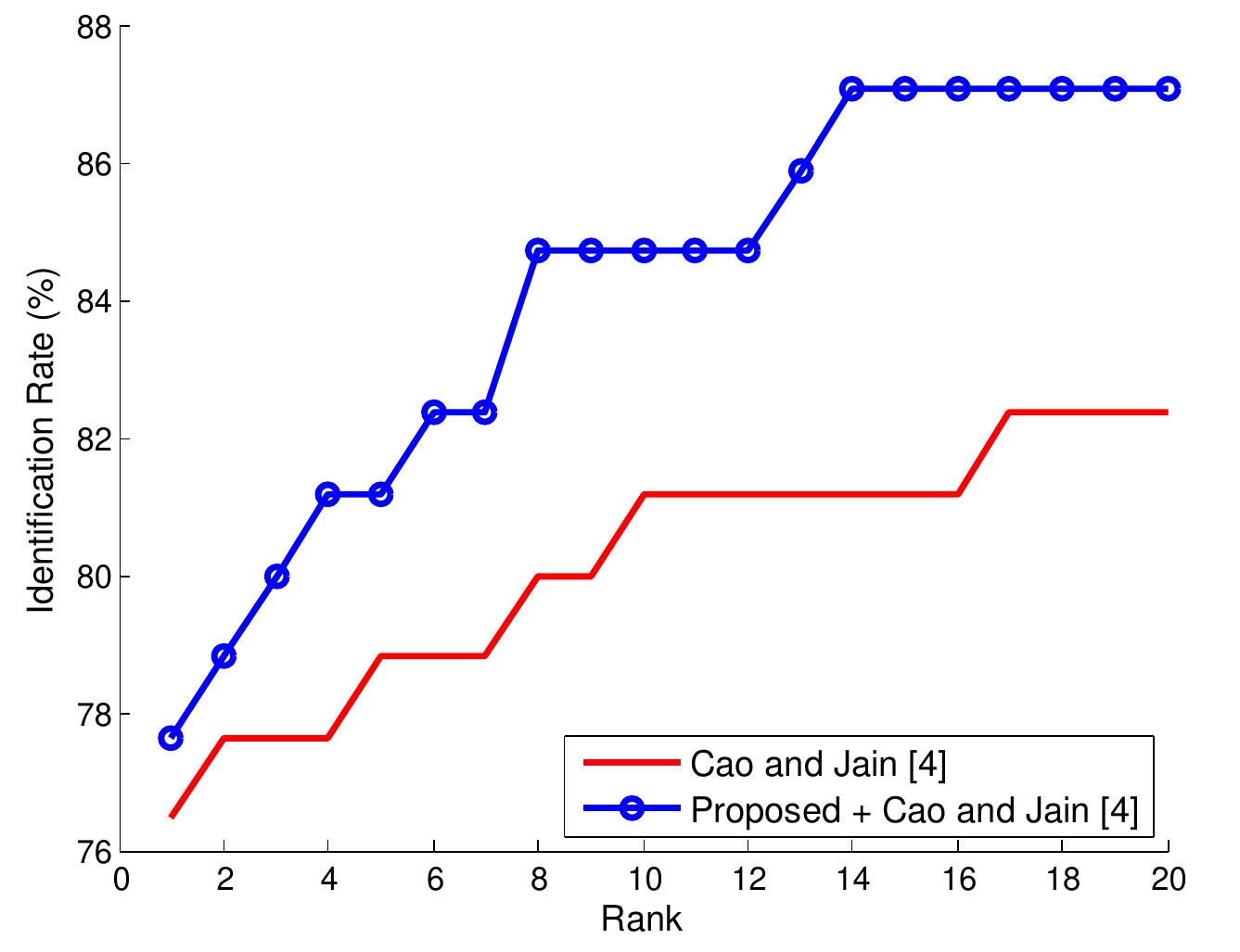}
	}   \hspace{ 1cm}
	\subfigure[]{
		\includegraphics[clip, trim=4cm 8.5cm 4cm 9cm, width=0.35\linewidth]{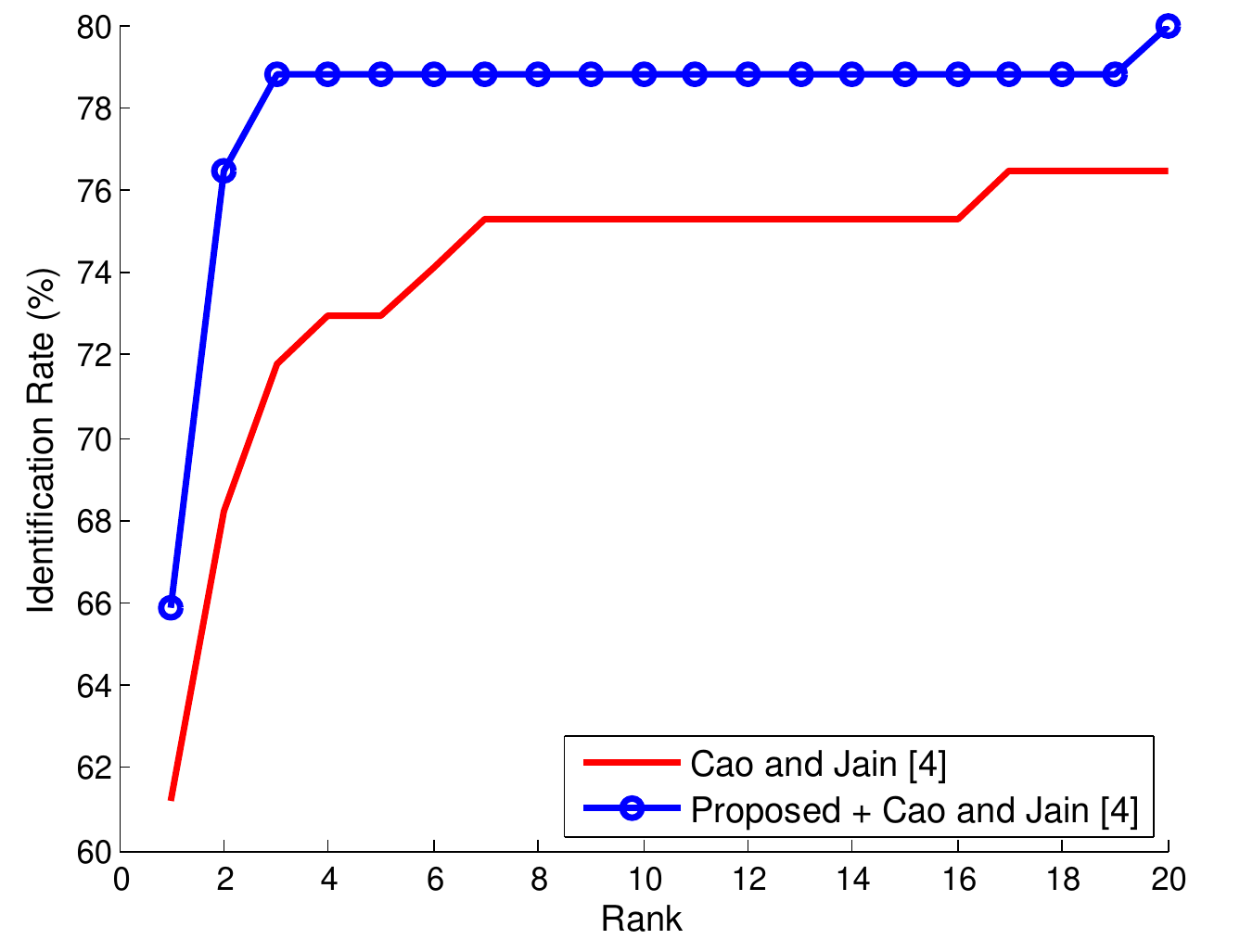}
	}           
	\caption{Cumulative Match Characteristic (CMC) curves of the three texture templates proposed here and their  fusion with three templates used in \cite{Cao2018PAMI} on (a) all 258 latents in NIST SD27, (b) subset of 88 good latents, (c) subset of 85 bad latents and (d) subset of 85 ugly latents.  Note that the scales of the y-axis in these four plots are different to show the differences between the two curves.}
	\label{fig:CMC}
	\vspace*{-5pt}
\end{figure*}

\begin{table}[htp]
\caption{Latent search accuracies under different scenarios for NIST SD27.  Reference database size is 10K. }
\vspace*{-10pt}
\begin{center}
\begin{tabular}{|p{1.9cm}|p{1.3cm}|p{1.cm}|p{1.cm}|p{1.1cm}|}
\hline
Input templates & descriptor length &rank-1 (\%)& rank-5 (\%)& rank-10 (\%) \\
\hline
Cao\&Jain \cite{Cao2018PAMI} & 384 & 59.30 & 70.16& 73.26 \\
\hline
\hline
$T_{e_1}$ & 192 & 68.22 & 73.64 & 74.81 \\
\hline
$T_{e_2}$& 192 & 66.67 & 72.48 & 74.42 \\
\hline
$T_{t}$ & 192 &  60.47 & 67.83 & 70.93   \\
\hline
$T_{e_1}$+$T_{e_2}$ & 192 & $\mathbf{70.93}$ & 74.81 & 77.91  \\
\hline
$T_{e_1}$+$T_{t}$ & 192 & $\mathbf{70.93}$ & 76.36 & 79.07   \\ 
\hline
$T_{e_2}$+$T_{t}$ & 192 &  67.05 & 75.19 & 77.13\\
\hline
$T_{e_1}$+$T_{e_2}$+$T_{t}$ & 192 & 70.16 & $\mathbf{76.74}$ & $\mathbf{81.40}$ \\
\hline
$T_{e_1}$ & 384 &69.38 & 75.58 & 77.13  \\
\hline
$T_{e_2}$ & 384 & 66.28 & 72.48 & 73.64 \\
\hline
$T_{t}$  & 384 & 58.91 & 66.28 & 69.77\\
\hline
$T_{e_1}$+$T_{e_2}$ & 384 & 70.16 & 75.58 & 78.29 \\
\hline
$T_{e_1}$+$T_{t}$  & 384 & 69.38 & 76.74 & 77.91 \\ 
\hline
$T_{e_2}$+$T_{t}$  & 384 &   67.83 & 74.03 & 75.97 \\
\hline
$T_{e_1}$+$T_{e_2}$+$T_{t}$ & 384 & $\mathbf{70.93 }$ & 75.97 & 78.68 \\

\hline
\end{tabular}
\end{center}
\label{tab:accuracy}
\vspace*{-10pt}
\end{table}%

\subsection{Texture template search}
In this section, we evaluate the search performance of the proposed three texture templates, i.e., $T_{e_1}$,  $T_{e_2}$ and $T_{t}$ extracted in section \ref{sec:templates}, as well as their fusion. The following seven scenarios are considered:
\begin{enumerate}
\itemsep0em 
\item $T_{e_1}$:  Texture template with enhanced image 1 (Fig. \ref{fig:input_images} (b)) for descriptor extraction;
\item $T_{e_2}$:  Texture template with enhanced image 2 (Fig. \ref{fig:input_images} (c)) for descriptor extraction;
\item $T_{t}$:  Texture template with texture image (Fig. \ref{fig:input_images} (d)) for descriptor extraction;
\item $T_{e_1}$ + $T_{e_2}$:  Score level fusion of $T_{e_1}$ and $T_{e_2}$;
\item $T_{e_1}$ + $T_{t}$:  Score level fusion of $T_{e_1}$ and $T_{t}$;
\item $T_{e_2}$ + $T_{t}$:  Score level fusion of $T_{e_2}$ and $T_{t}$;
\item $T_{e_1}$ + $T_{e_2}$ + $T_{t}$:   Score level fusion of $T_{e_1}$, $T_{e_2}$ and $T_{t}$.
\end{enumerate}
For each one of the above seven scenarios, all three descriptor lengths, i.e., $l_d$ = 96, 192 and 384, are considered.  The latent search accuracies of different scenarios at three different ranks  for  NIST SD27  against 10K reference fingerprints are shown in  Table \ref{tab:accuracy}. In addition,  the performance of texture template used in \cite{Cao2018PAMI} is also included for a comparison  (row 1 of Table \ref{tab:accuracy}). Note that the enhanced images used for $T_{e_1}$ extraction are the same as those in \cite{Cao2018PAMI}.  A descriptor length 96 performs much lower than the other two lengths (192 and 384) so its performance is not reported in Table \ref{tab:accuracy}.  The findings from Table \ref{tab:accuracy} can be summarized as follows: i)  among the three texture templates, $T_{e_1}$ performs the best for both descriptor lengths; ii) the performance of descriptor length of 192 is sufficiently  close to that of descriptor length 384 in different scenarios; iii) rank-1 accuracy of $T_{e_1}$ with descriptor length 192 is 8.98\% higher than that in \cite{Cao2018PAMI}; and iv) fusion of any two out of the three proposed texture templates is higher than that of any single template; the fusion of all three templates boosts the performance slightly  at rank-1 (for $l_d$=192) and rank-5 (for $l_d$=384).  Fusion of all three proposed templates with $l_d=192$ is preferred because of its smaller feature length  and  higher accuracy at rank-10. As examiners typically evaluate top 10 candidate matches to identify the source of a latent \cite{Busey2010}, this will ensure that the true mate is frequently available in the candidate list.

%descriptor length 192 and 284 are considered  while 
%The Cumulative Match Characteristic (CMC) curves of different descriptor lengths by matching 258 latents against 10K reference fingerprints are shown in Fig. \ref{fig:cmc_embedding_size}. In this experiment, we fix the value of stride, i.e., 16 pixels for both latents and rolled prints which are exactly the same as \cite{Cao2018PAMI}, for virtual minutiae extraction, and use the modified recognition algorithm in Section \ref{sec:matching} for matching. The performances of  descriptors with length 384 and 196  are close to each other and perform much better than that of the approach in \cite{Cao2018PAMI}, specifically, they are 10.1\% and 8.9\% higher at rank-1, while performance with descriptor length 96 is much lower than other two.  Overall, descriptor length 192 achieves  good balance between identification accuracy and feature length. 

\subsection{Fusion with minutiae templates}

 \begin{figure}[t] 
	\centering
	\subfigure[]{
		\includegraphics[width=0.4\linewidth]{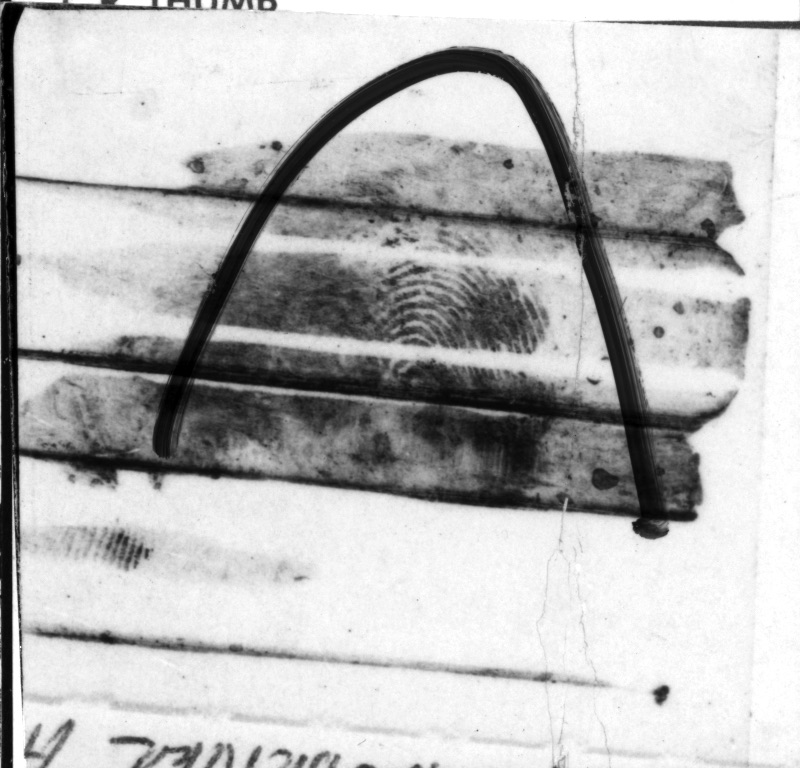}
	} \hspace{0.5cm}
	\subfigure[]{
		\includegraphics[ width=0.4\linewidth]{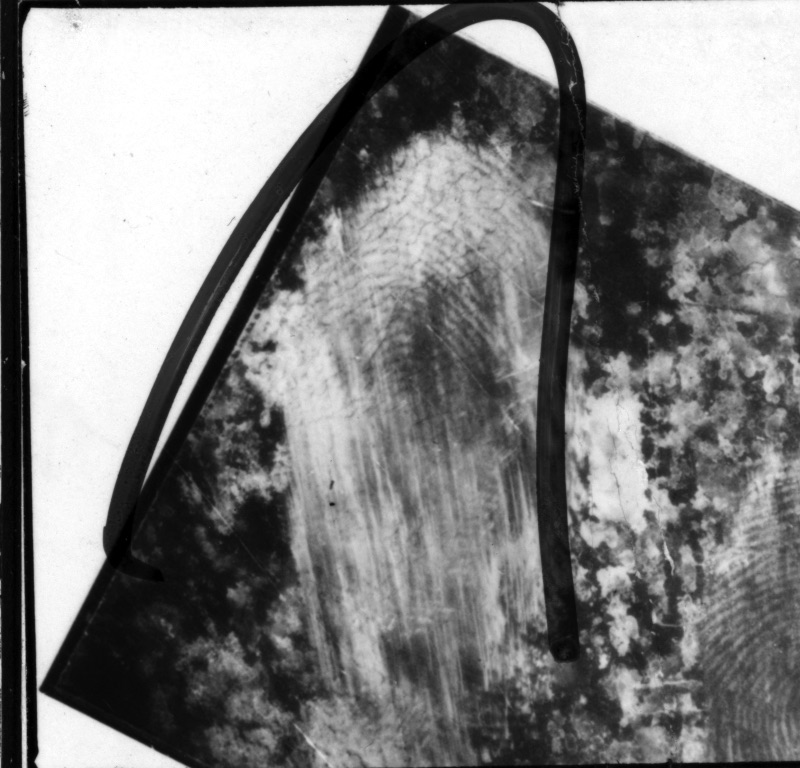}
	}     \hspace{1 cm}
	\caption{Two ugly latents whose true mates were not retrieved at rank 1 by the algorithm in \cite{Cao2018PAMI}. The use of proposed texture templates and their fusion with templates in \cite{Cao2018PAMI}  correctly retrieves their true mates at rank 1. }
	\label{fig:failure}
	\vspace*{-10pt}
\end{figure}

In order to determine if our new texture templates can boost the performance over the results in  \cite{Cao2018PAMI} we fuse the proposed three texture templates with the three templates used in \cite{Cao2018PAMI}. Fig. \ref{fig:CMC}  compares the Cumulative Match Characteristic (CMC) curves of the fusion scheme  on all 258 latents in NIST SD27 as well as subsets of latents of three different quality levels  (good, bad and ugly).  Plots in Fig. \ref{fig:CMC}  show the proposed three texture templates when fused with the three templates in \cite{Cao2018PAMI} can boost the overall performance by 2.7\% at rank-1 (from 75.6\% to 78.3\%). In particular,  the fusion of six templates  (three proposed + three from \cite{Cao2018PAMI})  improves the rank-1 accuracy by 4.7\% on the subset of ugly latents, some of the most challenging latents with an average of only 5 minutiae per latent. Fig. \ref{fig:failure} shows two ugly latents  whose true mates were not retrieved at rank-1 by the method in \cite{Cao2018PAMI}, but are now correctly retrieved at rank-1 with the introduction of three new texture templates.
%
%In this section, we fix the descriptor length at 192 and evaluate the balance between identification accuracy and stride size. Three stride size configurations are considered: i) 16 pixels for both latents and rolled prints, ii) 16 pixels for latents and 32 pixels for rolled prints, and iii) 32 pixels for both latents and rolled prints. The average number of virtual minutiae and the average time for feature extraction for different strides are shown in Table \ref{tab:statistics}.
%
%The CMC curves of these three scenarios as well as the approach in \cite{Cao2018PAMI} are shown in Fig. \ref{fig:stride}.  The performance drops as the stride increases but the matching speed is significantly improved. 

\subsection{Computation time}
The texture template matching algorithm was implemented in tensorflow and python and executed on a desktop with  i7-6700K CPU@4.00GHz, GTX 1080 Ti (GPU),  32 GB RAM and Linux operating system. The average computation time for comparing a latent texture template to a rolled texture template is 7.7ms  (single thread)  compared to 11.0 ms (24 threads) in \cite{Cao2018PAMI}. The average  times for extracting one proposed latent texture template and one proposed rolled texture template are 0.7s and 1.5s (GPU), respectively;  when fused with the three templates in \cite{Cao2018PAMI}  are 1.2s and 2.2s (24 threads).

% ------ % % % % % % ------
%\begin{figure}[t]
%    \centering
%     \subfigure[]{
%        \includegraphics[width=0.2\linewidth]{fig/Noise_1.eps}
%        }
%        \hspace{0.2 cm}
%    \subfigure[]{
%        \includegraphics[width=0.2\linewidth]{fig/Noise_2.eps}
%        }
%        \hspace{0.2 cm}
%    \subfigure[]{
%        \includegraphics[width=0.2\linewidth]{fig/Noise_3.eps}
%        }\hspace{0.2 cm}
%        \subfigure[]{
%    \includegraphics[width=0.2\linewidth]{fig/Noise_original.eps}
%    }\hspace{0.2 cm}
%    \subfigure[]{
%        \includegraphics[width=0.2\linewidth]{fig/Noise_texture.eps}
%        }        \hspace{0.2 cm}
%    \subfigure[]{
%        \includegraphics[width=0.2\linewidth]{fig/Noise_corrupted.eps}
%        }
%    \caption{Illustration of the proposed method for latent noise simulation  in training patches: (a)-(c) three
%    texture noise images generated by Eq. (\ref{eq:gabor}),
%    (d) a fingerprint patch, (e) a corrupted fingerprint patch with texture in (a)-(c), and
%    (f) around 30\% blocks in (e) are set to zero.  }
%      \label{fig:noisesimilation}
%\end{figure}
%------------------------------------------------------------------------
\section{Summary and future work}

Texture template is critical to improve the  search accuracy of latent fingerprints, especially for latents with small friction ridge area  and large background noise. We have proposed a set of three texture templates, defined as a set of  virtual minutiae along with their descriptors. Different virtual minutiae descriptors lead to different texture templates. The contributions of this paper are as follows. i) Use patches from original fingerprints and enhanced fingerprints to improve the distinctiveness of virtual  minutiae descriptors, ii) three different texture templates, and iii) a modified second-order graph matching. 
Identification results on NIST SD27 latent database demonstrate that the proposed texture templates when used alone can improve the rank-1 accuracy  by 8.9\% (from 59.3\% in \cite{Cao2018PAMI} to 68.2\%).   Our ongoing research includes i) improving ridge flow estimation, ii) using latent-rolled pairs for learning minutiae descriptors and similarities, and iii) using multicore processors to improve the search speed. 
\section*{Acknowledgement}
This research is based upon work supported in part by the Office of the Director of National Intelligence (ODNI), Intelligence Advanced Research Projects Activity (IARPA), via IARPA R\&D Contract No. 2018-18012900001. The views and conclusions contained herein are those of the authors and should not be interpreted as necessarily representing the official policies, either expressed or implied, of ODNI, IARPA, or the U.S. Government. The U.S. Government is authorized to reproduce and distribute reprints for governmental purposes notwithstanding any copyright annotation therein.
% 
%Orientation field estimation is a critical step to a robust
%lights-out latent fingerprint identification system. While
%dictionary based approaches have shown some success  for orientation
%field estimation, they have the following drawbacks: (1) the initial
%orientation field used in orientation patch dictionary construction
%itself is not reliable, (2) extending the ridge structure dictionary
%with large patch size results in too many dictionary elements and
%reduces their efficacy, (3) only a relatively small number of
%fingerprints have been used for training in dictionary based
%approaches, and (4) the dictionaries, typically learnt from high
%quality fingerprints, may not work well on poor quality latents. In
%this paper, we have proposed a ConvNet based fingerprint orientation
%field estimation algorithm by posing orientation field estimation in
%a latent patch as a classification task. Experimental results on
%NIST SD27 latent database demonstrate the superior performance of
%the proposed method. This can be attributed to: (1) the inherent
%power of ConvNet for learning a robust representation for
%classification, (2) a large number fingerprint patches with
%simulated noise used for training the ConvNet, and (3) a
%preprocessing method to enhance potential ridge structure and
%suppress texture noise.
%
%
%Future work will include evaluating the effect of (1) orientation
%pattern size, and (2) noise level added to the training patches, and
%(3) estimating the ridge frequency field using the same ConvNet
%based  framework.

{\footnotesize
\bibliographystyle{ieee}
\bibliography{FM_Ref}
}

\end{document}